\def\year{2022}\relax
\documentclass[letterpaper]{article} % DO NOT CHANGE THIS
\usepackage{aaai22}  % DO NOT CHANGE THIS
\usepackage{times}  % DO NOT CHANGE THIS
\usepackage{helvet}  % DO NOT CHANGE THIS
\usepackage{courier}  % DO NOT CHANGE THIS
\usepackage[hyphens]{url}  % DO NOT CHANGE THIS
\usepackage{graphicx} % DO NOT CHANGE THIS
\usepackage{subcaption} % zpz
\usepackage{natbib}  % DO NOT CHANGE THIS AND DO NOT ADD ANY OPTIONS TO IT
\usepackage{caption} % DO NOT CHANGE THIS AND DO NOT ADD ANY OPTIONS TO IT
\DeclareCaptionStyle{ruled}{labelfont=normalfont,labelsep=colon,strut=off} % DO NOT CHANGE THIS
\usepackage{amsfonts} %zpz
\usepackage{diagbox} %zpz
\usepackage{bbding} %zpz
\usepackage{multirow} % zpz use, should check
\usepackage[colorlinks,linkcolor=blue]{hyperref}
\title{LGD: Label-guided Self-distillation for Object Detection}

\author{
    Peizhen Zhang,\equalcontrib\textsuperscript{\rm 1}
    Zijian Kang,\equalcontrib\textsuperscript{\rm 2}
    Tong Yang,\textsuperscript{\rm 1}
    Xiangyu Zhang,\thanks{Corresponding author.}\textsuperscript{\rm 1}\\
    Nanning Zheng,\textsuperscript{\rm 2}
    Jian Sun \textsuperscript{\rm 1}}
\affiliations{
    \textsuperscript{\rm 1}MEGVII Technology,\\
    \textsuperscript{\rm 2}Xi'an Jiaotong University\\
    \{zhangpeizhen, yangtong, zhangxiangyu, sunjian\}@megvii.com,\\
    kzj123@stu.xjtu.edu.cn,
    nnzheng@mail.xjtu.edu.cn
}

\begin{document}

%\linenumbers

\maketitle

\begin{abstract}
In this paper, we propose the first self-distillation framework for general object detection, termed LGD (\textbf{L}abel-\textbf{G}uided self-\textbf{D}istillation). Previous studies rely on a strong pretrained teacher to provide instructive knowledge that could be unavailable in real-world scenarios. Instead, we generate an instructive knowledge based only on student representations and regular labels. Our framework includes sparse label-appearance encoder, inter-object relation adapter and intra-object knowledge mapper that jointly form an implicit teacher at training phase, dynamically dependent on labels and evolving student representations. They are trained end-to-end with detector and discarded in inference. Experimentally, LGD obtains decent results on various detectors, datasets, and extensive tasks like instance segmentation. For example in MS-COCO dataset, LGD improves RetinaNet with ResNet-50 under $2\times$ single-scale training from 36.2\% to 39.0\% mAP (+ \textbf{2.8}\%). It boosts much stronger detectors like FCOS with ResNeXt-101 DCN v2 under $2\times$ multi-scale training from 46.1\% to 47.9\% (+ \textbf{1.8}\%). Compared with a classical teacher-based method FGFI, LGD not only performs better without requiring pretrained teacher but also reduces \textbf{51}\% training cost beyond inherent student learning. Codes are available at \href{https://github.com/megvii-research/LGD}{https://github.com/megvii-research/LGD}.
\end{abstract}

\section{Introduction}
\label{sec:intro}
Knowledge distillation (KD) \cite{fitnet,kd} is initially proposed for image classification and obtains impressive results. Typically, it is about transferring instructive knowledge from a pretrained model (teacher) to a smaller one (student). Recently, KD applied to the fundamental object detection task, has aroused researchers' interests \cite{li2017mimicking,wei2018quantization,wang2019distilling,zhang2020prime,dai2021general,guo2021distilling,zhang2021improve,yao2021gdetkd}. 
Existing works achieve respectable performance but the choice of teacher is sophisticated and inconsistent among them. 
One common ground is that they all require a heavy pretrained teacher as it is discovered by recent works \cite{zhang2021improve,yao2021gdetkd} that distillation efficacy could be enhanced with stronger teachers. Yet the pursuit for an ideal teacher could scarcely be satisfied in real-world applications, since it might take tons of efforts on trial and error \cite{peng2020cream}. 
Instead, the issue that ``\textbf{\textit{KD for generic detection without pretrained teacher}}" is barely investigated.

\begin{figure}[t]
    \centering
    \includegraphics[width=0.9\linewidth]{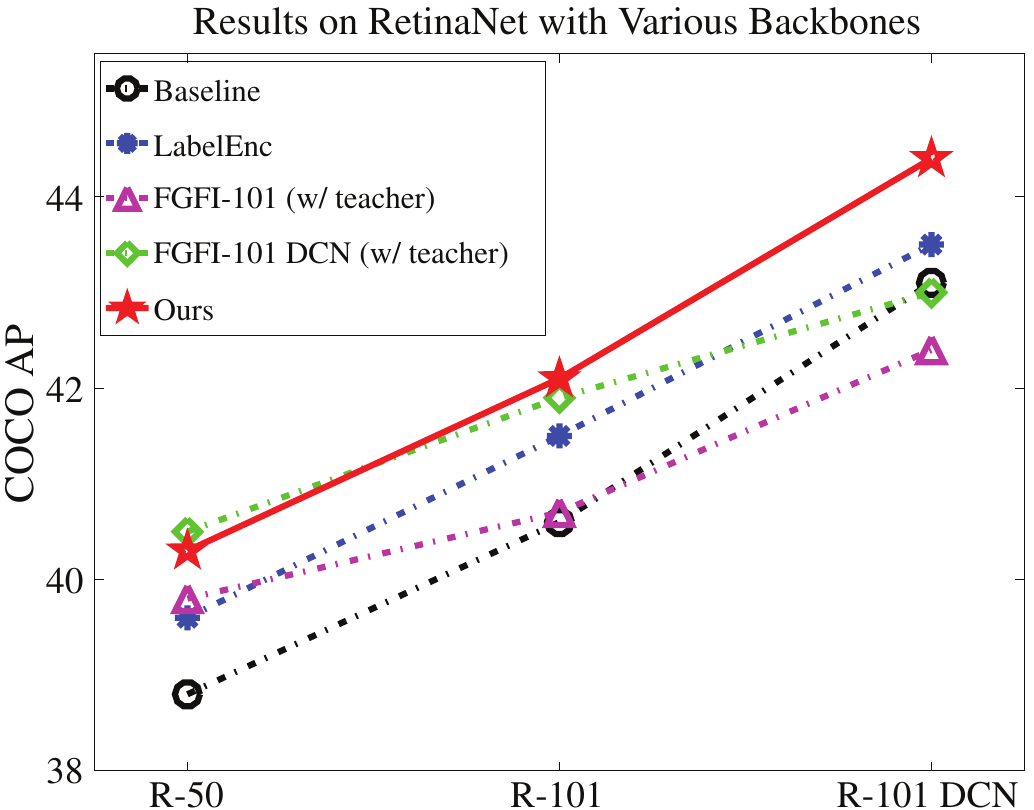}
    \centering
    \caption{Results trending on RetinaNet 2$\times$ $ms$ with backbones R-\{50, 101, 101 DCN\} respectively. FGFI-\{101, 101 DCN\} denote FGFI method using RetinaNet 2$\times$ $ms$ with R-101 and R-101 DCN as teachers, respectively.}
    \label{fig:distill-50-101-dcnv2-retinanet}
\end{figure}

\begin{figure*}[htbp]
    \centering
    \includegraphics[width=1.0\linewidth]{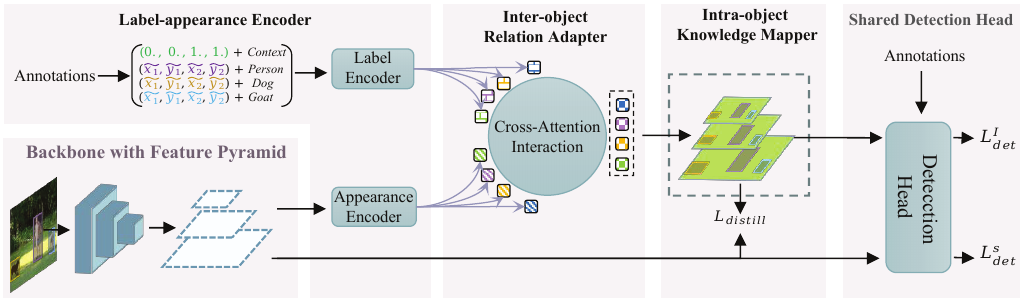}
    \caption{The proposed framework contains three modules: (1) Label-appearance encoder, (2) Inter-object relation adapter and (3) Intra-object knowledge mapper. For brevity, we omit the pyramid level indications which will be elaborated in Section~\ref{sec:method}. $L_{det}^{I}$\,/\,$L_{det}^{S}$ denote detection losses upon instructive\,/\,student representations and $L_{distill}$ is the distillation loss. We denote by $(\widetilde{x_1}, \widetilde{y_1}, \widetilde{x_2}, \widetilde{y_2})$ the ground-truth box location normalized by image size that $(0., 0., 1., 1.)$ refers to an entire context box.}
    \label{fig:framework}
\end{figure*}

To alleviate the pretrained teacher dependence, teacher-free schemes are proposed like (a) \textit{self-distillation}, (b) \textit{collaborative learning} and (c) \textit{label regularization}, where instructive knowledge could be cross-layer features \cite{zhang2019your}, competitive counterparts \cite{zhang2018deep} and modulated label distribution \cite{yuan2020revisiting}, etc. However, these methods are designed for classification and are inapplicable to detection since the latter has to handle multiple objects with different locations and categories but singe image classification. Lately, LabelEnc \cite{hao2020labelenc} extends traditional label regularization by introducing location-category modeling with an isolated network. It produces label representations with which the student features are supervised. Though it obtains impressive results, we find the improvement saturates (Figure~\ref{fig:horizontal_comparison}) as detector grows stronger, \textit{e.g.}, with larger backbones and multi-scale training.
We conjecture this is because labels themselves describe only object-wise categories and locations, without considering the inter-object relationship which is also important \cite{hu2018relation,cai2019exploring}. For detectors with limited capacity, LabelEnc provides strong complementary supervision, albeit without relation information. For stronger detectors which are able to extract abundant object-wise hints from default supervision, using LabelEnc becomes less beneficial or even detrimental (see the leftmost figure in Figure~\ref{fig:horizontal_comparison}). This might result from semantic discrepancy by heterogeneous input (image \textit{vs.} label) and isolated modeling.

Motivated by this, we propose \textbf{L}abel-\textbf{G}uided self-\textbf{D}istillation (LGD), a new teacher-free method for object detection as shown in Figure~\ref{fig:framework}.
In LGD, we devise an inter-object relation adapter and an intra-object knowledge mapper to collaboratively model the relation in forming instructive knowledge. The relation adapter computes interacted embeddings by a cross-attention interaction. 
Specifically, the interacted embedding of each object is calculated by first measuring the cross-modal similarity between its appearance embedding and every label embedding upon which a weighted-aggregation is then performed.
The knowledge mapper maps the interacted embeddings onto feature map space as final instructive knowledge, considering intra-object representation consistency and localization heuristics. Owing to the above relation modeling, the final instructive knowledge is naturally adapted to the student representations, facilitating effective distillation for strong student detectors and semantic discrepancy mitigation.
Beyond efficacy, our method is also efficient, it does not rely on a strong convolution network as teacher because we adopt efficient instance-wise embeddings design. The above efficient design allows LGD to train jointly with the student, simplify the pipeline, and reduce training cost (Table~\ref{tab:train_cost}). During inference, only student detector is kept, bringing no extra cost. In short, our contributions are three-fold:

\begin{enumerate}
    \item We propose a new self-distillation framework for general object detection. Unlike previous methods that use a convolution network as teacher, LGD generates instructive knowledge on-the-fly without pretrained teacher and improves the detection quality under limited training cost.
    
    \item We introduce inter-and-intra relation to model a new instructive knowledge, rather than simply extract existent relation from student and teacher for distillation. 
    
    \item The proposed method outperforms previous teacher-free SOTA with higher upper limit and is better than classical teacher-based method FGFI in strong student settings. Beyond inherent student learning, it saves \textbf{51}\% training time against the classical teacher-based distillation.
\end{enumerate}

\section{Related Work}

\subsection{Detection KD with Pretrained Teachers}
\label{subsec:related_work:KD_for_det}
Unlike classification, knowledge transfer for object detection is more challenging. In detection, models are asked to predict multiple instances with diversified categories distributed at different locations in the image. \cite{li2017mimicking} proposed Mimic to distill activations within the region proposals predicted by RPN\,\cite{ren2015faster}. \cite{chen2017learning} introduced weighted cross-entropy and bounded regression loss for enhancing the performance. To further exploit the context information of the distilling regions around the objects, \cite{wang2019distilling} extended the ground-truth box regions by anchor-assigned ones. For learning adapted sampling weight for different knowledge, \cite{zhang2020prime} proposed PAD with uncertainty modeling. Besides intermediate feature hints, \cite{dai2021general} involved the prediction map distillation obeying the assignment rules and relation distillation\,\cite{park2019relational} upon their defined general instances. Instead of focusing on foreground regions only, \cite{guo2021distilling} decoupled the fore/back-ground knowledge transfer. To facilitate region-agnostic distillation, \cite{zhang2021improve} proposed feature-based knowledge transfer by spatial-channel-wise attention. To resolve the feature resolution mismatching in cross-layer distillation and mitigate the misaligned label assignment, \cite{yao2021gdetkd} introduced G-DetKD. Above methods mainly conducted feature-based distillation which is followed in this work. Whereas, they are designed for settings with strong pretrained teachers that could be unavailable or unaffordable in real-world scenarios. Recently, \cite{huang2020comprehensive} proposed self-distillation for weakly supervised detection but the setting is much different from generic object detection.

\subsection{Teacher-free Methods}
\label{subsec:related_work:Teacher_free_method}
Beyond traditional KD with pretrained teacher, there are teacher-free schemes that could be divided into three categories: (1) self-distillation (2) collaborative learning and (3) label regularization. (1) self-distillation excavates instructive knowledge from model itself. For instance, \cite{yang2019snapshot, kim2020self} used previously saved snapshots as teachers. In \cite{zhang2019your}, network was divided into sections that deeper layers were used to teach the shallower ones. In MetaDistiller\,\cite{liu2020metadistiller}, the knowledge stemmed from one-step predictions. (2) Collaborative learning involves multiple students to boost each other. \cite{zhang2018deep} proposed deep mutual learning (DML) where student networks with identical architecture learned collaboratively. \cite{lan2018knowledge} proposed ONE by considering ensemble learning in branch-granularity. In KDCL\,\cite{guo2020online}, predictions were fused together as instructive knowledge. Likewise in \cite{chen2020online}, ensemble logits of multiple students were aggregated to distill another. \cite{furlanello2018born} proposed Born-Again Network (BAN) that leveraged information from last generations to distill the next. (3) For label regularization, \cite{yuan2020revisiting} proposed tf-KD for regularized label distribution beyond label smoothing\,\cite{szegedy2016rethinking}. However, above methods were designed for classification only.

Recently, there have been newly-built label regularization methods\,\cite{mostajabi2018regularizing,hao2020labelenc} using an isolated network to explicitly model labels as features for supervision,\textit{w.r.t.} semantic segmentation and detection. They obtained impressive results. In \cite{hao2020labelenc}, dense color maps with category and location information were constructed and fed into an auto-encoder-like network to fetch label representations. However, they considered each object modeling separately which was sub-optimal. Instead, we propose to generate instructive knowledge by inter-object and intra-object relation modeling to form a self-distillation scheme with higher upper limit.

\section{Method}
\label{sec:method}
As shown in Fig.~\ref{fig:framework}, we illustrate the modules in LGD as follows: (1) An encoder that computes label and appearance embeddings. (2) An inter-object relation adapter that generates interacted embeddings given label and appearance embeddings of objects. (3) An intra-object knowledge mapper that back-projects interacted embeddings onto feature map space to obtain instructive knowledge for distillation.

\subsection{Label-appearance Encoder}
\label{subsec:encoder}

\noindent\textbf{(1) Label Encoding:}
For each object, we concatenate its normalized ground-truth box $(\widetilde{x_1}, \widetilde{y_1}, \widetilde{x_2}, \widetilde{y_2})$ and one-hot category vector to obtain a descriptor. The object-wise descriptors are passed into a label encoding module for refined label embeddings $\mathcal{L} = \{\mathbf{l}_{i} \in \mathbb{R}^C \}_{i=0}^{N}$, where $i$ indicates object index, $C=256$ is the intermediate feature dimension, and $N$ is the object number. $i=0$ indexes the context object. To introduce basic relation modeling among label descriptors and maintain a permutation-invariant property, we adopt the classical PointNet \cite{qi2017pointnet} as the label encoding module. It processes the descriptors by a multi-layer perceptron\,\cite{friedman2001elements} with local-global modeling by a spatial transformer network\,\cite{jaderberg2015spatial}. Also, the label descriptors are similar to point set that is accustomed to PointNet (bounding boxes could be viewed as points in 4-dimensional Cartesian space). Empirically, using PointNet as encoder behaves slightly better than MLP or transformer encoder\,\cite{vaswani2017attention} (Table~\ref{tab:encoder_ablation}). We further replace the BatchNorm\,\cite{ioffe2015batch} with LayerNorm\,\cite{ba2016layer} to adapt the small-batch detection setting. Notably, the above 1D object-wise label encoding manner is more efficient than that in LabelEnc. The LabelEnc constructs an ad-hoc color map $\in \mathbb{R}^{H\times W\times K}$ to describe labels where $(H, W)$ and $K$ are input resolution and object category number respectively ($HWK \gg C$). The color map is processed by an extra CNN and pyramid network for 2D pixel-wise representations $\mathcal{L'} = \{{\mathbf{l}'}_{i} \in \mathbb{R}^{H_p \times W_p \times C}, 1\leq p \leq P\}$. $P$ refers to the number of pyramid scales\,\cite{lin2017feature} that $(H_p, W_p)$ denotes feature map resolution at scale $p$. 
\\

\noindent\textbf{(2) Appearance Encoding:} Beyond label encoding, we retrieve compact appearance embeddings from feature pyramid of student detector that contains appearance feature of perceived objects. 
We adopt a handy mask pooling to extract object-wise embeddings from the feature maps. Specifically, we pre-compute the object-wise masks: $\mathcal{M}=\{\mathrm{m}_i\}_{i=1}^{N} \bigcup \{\mathrm{m}_{0}\}$ at input level for total $N$ objects and a virtual context object with location $(0., 0., 1., 1.)$ covering the entire image. For each object $i$ ($0\leq i \leq N$), $\mathrm{m}_i \in \mathbb{R}^{H \times W}$ is a binary matrix whose values are set as $1$ inside the ground-truth region and $0$ otherwise. The mask pooling is conducted concurrently for all pyramid levels, at each of which, object-wise masks at input level are down-scaled to corresponding resolution to become scale-specific ones. At $p$-th scale, the appearance embedding $\mathbf{a}_i \in \mathbb{R}^{C}$ is obtained by calculating channel-broadcasted Hadamard product between the projected feature map $\mathcal{F}_{proj}(X_p) \in \mathbb{R}^{H_p\times W_p \times C}$ and down-scaled object mask $ \in \mathbb{R}^{H_p\times W_p}$, followed by global sum pooling. $\mathcal{F}_{proj}(\cdot)$ is a single $3\times 3$ \textit{conv} layer. Thus, we collect appearance embeddings: $\mathcal{A}_p = \{\mathbf{a}_i \in \mathbb{R}^{C}\}_{i=0}^{N}$ for each object at level $p$ (likewise for the other levels).

\subsection{Inter-object Relation Adapter}
\label{subsec:adapter}

Given label and appearance embeddings, we formulate the inter-object relation adaption by a cross-attention process. In Fig.~\ref{fig:framework}, this process is executed at every student appearance pyramid scale to retrieve the interacted embeddings. We omit the pyramid scale subscript below for brevity.

During the cross attention, a sequence of key and query tokens are leveraged in calculating KQ-attention relation for aggregating value to obtain attention outputs. For achieving the label-guided information adaption, we exploit the appearance embeddings $\mathcal{A}$ at current scale as query, and the scale-invariant label embeddings $\mathcal{L}$ as key and value. The attention scheme measures the correlation between lower-level structural appearance information and higher-level label semantics among objects then reassembles the informative label embeddings for dynamic adaption.

Before conducting attention, the query, key, and value are transformed by linear layers $f_{\mathcal{Q}}$, $f_{\mathcal{K}}$ and $f_{\mathcal{V}}$, respectively. We then computed the interacted embeddings $\mathbf{u}_i\in \mathbb{R}^C$ for $i$-th object by weighting each transformed label embedding $f_{\mathcal{V}}(\mathbf{l}_j)$ by label-appearance correlation factor $w_{ij}$.
\begin{equation}
     \mathbf{u}_{i} = \sum_{j=0}^{N}w_{ij}f_{\mathcal{V}}(\mathbf{l}_j)
\end{equation}
$w_{ij}$ is calculated by a scaled dot-product between $i$-th appearance embeddings $\mathbf{a}_i$ and $j$-th label embeddings $\mathbf{l}_j$ followed by a softmax operation:
\begin{equation}
\label{eq:w}
    w_{ij} = \frac{{\rm exp}\left( f_{\mathcal{Q}}(\mathbf{a}_{i}) \cdot f_{\mathcal{K}}(\mathbf{l}_{j})/\tau \right)}{\sum_{k=0}^{N}{{\rm exp}\left( f_{\mathcal{Q}}(\mathbf{a}_{i}) \cdot f_{\mathcal{K}}(\mathbf{l}_{k})/\tau \right)}}
\end{equation}
where $\cdot$ is the notation for inner product and $\tau = \sqrt{C}$ is the denominator for variance rectification \,\cite{vaswani2017attention}.

Specifically, for more robust attention modeling, the paradigm actually involves $T$ set of concurrent operations termed heads to obtain partial interacted embeddings in parallel. By concatenating the partial interacted embeddings from all heads and applying a linear projection $f_{\mathcal{P}}$, we obtain interacted embeddings $\mathbf{E} = \{\mathbf{e}_{i} \in \mathbb{R}^{C}\}_{i=0}^{N}$ for all objects: 
\begin{equation}
\label{eq:eik}
     \mathbf{e}_{i}=f_{\mathcal{P}}([\mathbf{u}^{1}_{i};\mathbf{u}^{2}_{i};\ldots;\mathbf{u}^{T}_{i}])
\end{equation}
where [;] denotes the concatenation operator that combines the partial embeddings along the channel dimension. The resulting embeddings are also scale-sensitive as the appearance embeddings. As aforementioned, we obtain interacted embeddings across scales by iterating over all feature scales.

Technically, above computation is accomplished by means of multi-head self attention (MHSA)\,\cite{vaswani2017attention}. Note that our framework is decoupled to the specific choice. As will be shown in this paper, LGD shows the efficacy even with the naive transformer. It is likely to perform even better by using advanced variants like focal transformer\,\cite{yang2021focal} but that is beyond the scope.

\subsection{Intra-object Knowledge Mapper}

To make the 1D interacted embeddings applicable to widely-used intermediate feature distillation\,\cite{li2017mimicking,wang2019distilling} for detection, we map the interacted embeddings onto 2D feature map space to fetch instructive knowledge. Naturally, for each pyramid scale $p, (1\leq p \leq P)$, the resolutions of resulting maps are confined to be identical with corresponding student feature maps.

Intuitively, since spatial topology is not maintained in label encoding for compact representations (Sec.~\ref{subsec:encoder}), it is important to recover the localization information for each object to achieve alignment in geometric perspective. Naturally, object bounding box regions serve as good heuristics. We fill each object-binding interacted embedding within its corresponding ground-truth box region on a zero-initialized feature map.
In practice, for each object $i$, we acquire its feature map at $p$-th scale by calculating matrix multiplication between the vectorized object mask $\mathbf{m}_{i} \in \mathbb{R}^{H_pW_p\times 1}$ and the projected, interacted embedding $\mathbf{e}_i$. All these object-wise maps are added up to a unified one followed by a refinement module $\mathcal{F}_{ref}(\cdot)$ to form the instructive knowledge:
\begin{equation}
\label{eq:projection}
     {X_p}^{\mathcal{I}} = \mathcal{F}_{ref}{\left[\mathbf{m}_0 \mathcal{F}^\top_{ctx}(\mathbf{e}_0)+ \mathcal{G}\left(\sum_{i=1}^{N}{\mathbf{m}_i \mathcal{F}^\top_{inst}(\mathbf{e}_i)}\right) \right]}
\end{equation}
where $\mathcal{F}^\top_{ctx}(\mathbf{e}_0)$ and $\mathcal{F}^\top_{inst}(\mathbf{e}_i)\in \mathbb{R} ^ {1 \times C}, (1\leq i \leq N)$ are the transposes of projected context and normal object interacted embeddings, respectively. Both $\mathcal{F}_{ctx}(\cdot)$ and $\mathcal{F}_{inst}(\cdot)$ are single \textit{fc} layers. $\mathcal{G}(\cdot)$ is a single $3\times 3$ \textit{conv} layer. $\mathcal{F}_{ref}(\cdot)$ starts with a \textit{relu} followed by three $3\times 3$ \textit{conv} layers. Thus, we collect the instructive knowledge $\mathcal{X}^\mathcal{I}=\{ {X_p}^{\mathcal{I}} \in \mathbb{R}^{H_p\times W_p \times C}\}_{p=1}^P$ at all scales.

Beyond applicability consideration, the above mapping implies a spirit of \textit{intra-object} regularization \cite{yun2020regularizing,law2018cornernet,chen2020reppointsv2} which enforces activation neurons inside the same foreground region on student appearance representations to be close (through subsequent distillation in Equation~\ref{eq:mse}). Moreover, these instructive representations will be supervised with detection loss for ensuring the representation capability (Equation~\ref{eq:det_loss}).

Before distillation, an adaption head $\mathcal{F}_{adapt}(\cdot)$ is used to adapt student representations, following FitNet. We conduct knowledge transfer between the instructive representations ${X_p}^{\mathcal{I}}$ and the adapted student features $X^{\mathcal{S}}_p=\mathcal{F}_{adapt}(X_{p})$ at each feature scale. We adopt InstanceNorm\,\cite{ulyanov2016instance} to eliminate the appearance and label style information for both feature maps followed by a Mean-Square-Error (MSE):

\begin{equation}
\label{eq:mse}
    L_{distill} = \frac{1}{N_{total}} \sum_{p=1}^{P} { \left \Vert X_{p}^{\mathcal{S}}-X_{p}^{\mathcal{I}} \right \Vert^2}
\end{equation}
where $P$ is the total number of pyramid levels, and $N_{total} = \sum_{p=1}^{P}H_pW_pC$ indicates the total size of the feature pyramid tensors. As gradient stopping technique suggested in previous studies \cite{hao2020labelenc, hoffman2016learning}, we detach instructive representations $\mathcal{X}^\mathcal{I}$ when calculating distillation loss to avoid model collapse.

Besides the distillation loss and detection loss for optimizing student detector, we further ensure the instructive representation quality and consistency with student representations by sharing the detection head for supervision. The overall detection loss is shown below:
\begin{equation}
\label{eq:det_loss}
    L_{det} = L_{det}^\mathcal{S}(\mathcal{H}(\mathcal{X}), \mathcal{Y}) + L_{det}^\mathcal{I}(\mathcal{H}(\mathcal{X}^{\mathcal{I}}), \mathcal{Y})
\end{equation}
where $\mathcal{X}$/$\mathcal{X}^\mathcal{I}$ denote student\,/\,instructive representations across scales. $L_{det}^\mathcal{S/I}$ denotes the detection loss (classification and regression) upon them. $\mathcal{H}(\cdot)$ refers to the detection head. $\mathcal{Y}$ stands for the label set (boxes and categories). In summary, the total training objective is:
\begin{equation}
\label{eq:tot_loss}
    L_{total} = L_{det} + \lambda L_{distill}
\end{equation}
where $\lambda$ is a trade-off for distillation term and we simply adopt $\lambda=1$ throughout all experiments. For stable training, the distillation starts in 30k iterations since it could be detrimental when the instructive knowledge is optimized insufficiently \cite{hao2020labelenc, liu2020metadistiller}. The student detector backbone is frozen in early $10k$ iterations under $1\times$ training schedule and $20k$ for $2\times$ training schedule.

\begin{figure*}[htbp]
    \centering
    \begin{subfigure}[t]{0.33\linewidth}
         \centering
         \includegraphics[width=\textwidth]{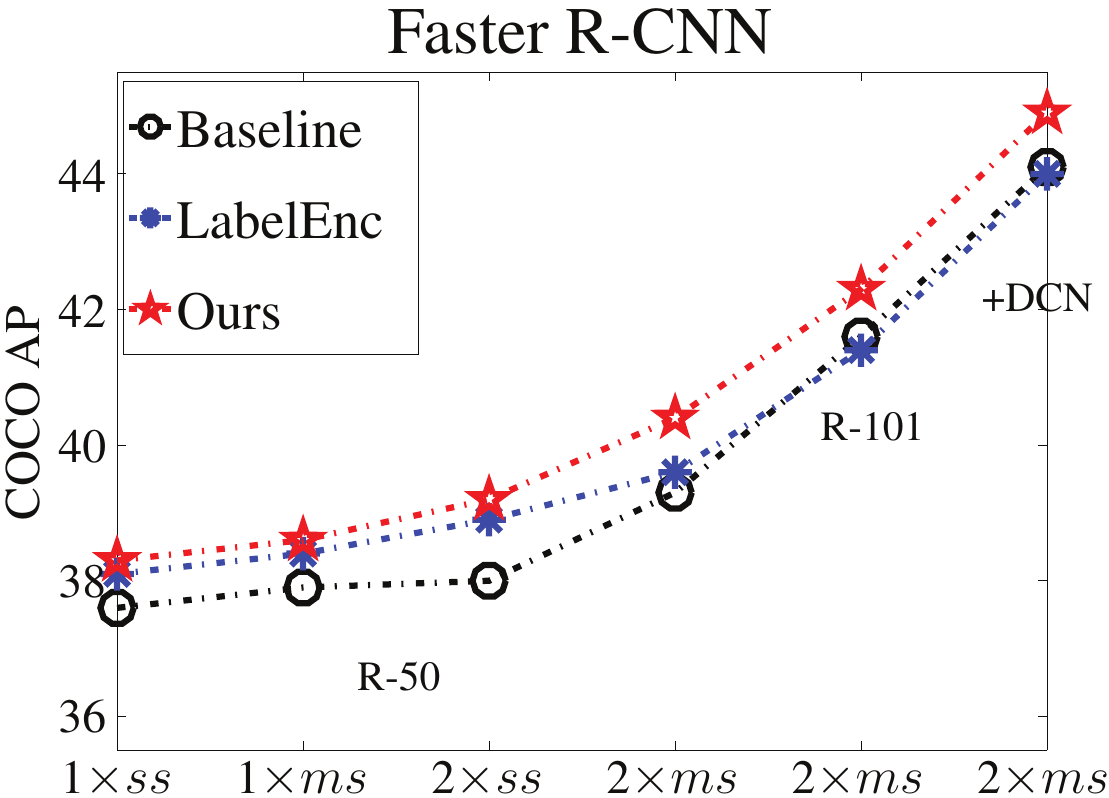}
         \label{fig:auxtask1}
     \end{subfigure}
     \begin{subfigure}[t]{0.33\linewidth}
         \centering
         \includegraphics[width=\textwidth]{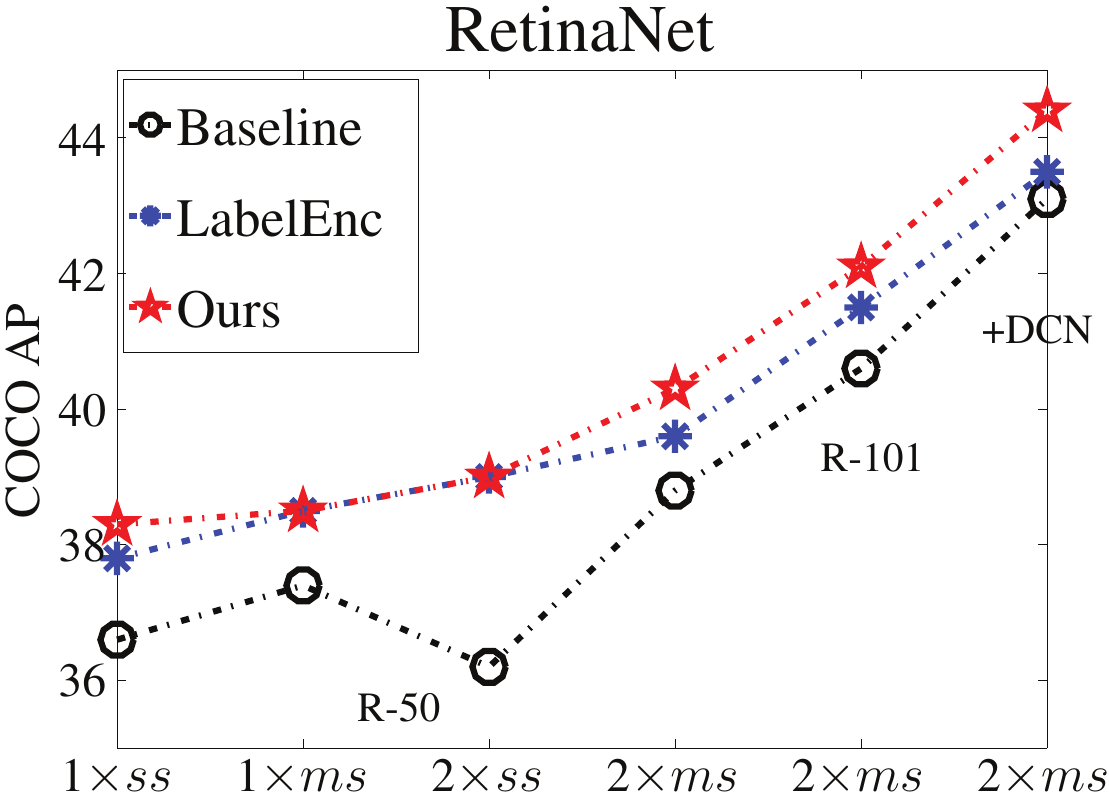}
         %\caption{}
         \label{fig:auxtask2}
     \end{subfigure}
     \begin{subfigure}[t]{0.33\linewidth}
         \centering
         \includegraphics[width=\textwidth]{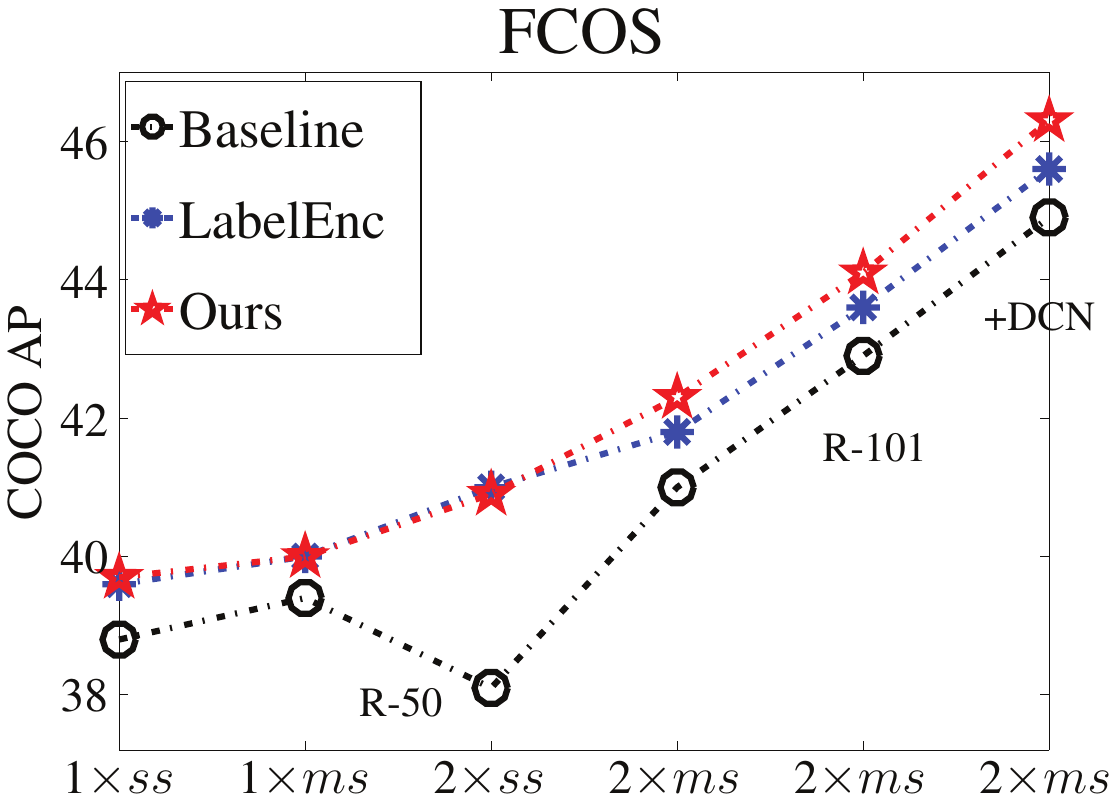}
         %\caption{}
         \label{fig:auxtask3}
     \end{subfigure}
    \caption{Result tendency as detector grows stronger on three typical detectors by LabelEnc and ours. In each sub-figure, there are six settings from left to right: R-50-\{1$\times$ $ss$, 1$\times$ $ms$, 2$\times$ $ss$, 2$\times$ $ms$\} $\rightarrow$ R-101-2$\times$ $ms$ $\rightarrow$ R-101 DCN-2$\times$ $ms$.}
    \label{fig:horizontal_comparison}
\end{figure*}

\section{Experiment}
\label{sec:expr}

\subsection{Experiments Setup}
The proposed framework is built upon Detectron2 \cite{wu2019detectron2}. Experiments are run with batch size 16 on 8 GPUs. Inputs are resized such that shorter sides are no more than 800 pixels. We use SGD optimizer with $0.9$ momentum and $10^{-4}$ weight decay. The multi-head attention in inter-object relation adapter uses $T=8$ heads following common practice. For brevity, we denote by R-50, R-101 and R-101 DCN for ResNet-50, ResNet-101 and ResNet-101 with deformable convolutions v2 \cite{zhu2019deformable}.
Main experiments are validated on MS-COCO \cite{lin2014microsoft} dataset that we also testify on others: Pascal VOC \cite{Everingham10} and CrowdHuman \cite{shao2018crowdhuman}.

\textbf{MS-COCO} is a challenging object detection dataset with 80 categories. Mean average precision (AP) is used as the major metric. Following common protocol\,\cite{he2019rethinking}, we use the \textit{trainval-115k} and \textit{minival-5k} subsets \textit{w.r.t.} training and evaluation. We denote by 1$\times$ the training for 90k iterations where learning rate is divided by 10 at 60k and 80k iterations. By analogy, 2$\times$ denotes 180k of iterations with milestones at 120k and 160k. We term the single and multi-scale training by $ss$ and $ms$ for short.

\textbf{Pascal VOC} is a dataset with 20 classes. The union of \textit{trainval-2007} and \textit{trainval-2012} subsets are used for training, leaving \textit{test-2007} for validation. We report mAP and AP50/75 (AP with overlapping threshold $0.5$/$0.75$). Models are trained for $24k$ iterations with milestones at $18k$ and $22k$. 

\textbf{CrowdHuman} is the largest crowd pedestrian detection dataset, containing 23 people per image. It includes 15k and 4370 images \textit{w.r.t.} training and validation. The major metric is \textit{average log miss rate over false positives per image} (termed mMR, lower is better). Models are trained for 30 epochs with learning rate decayed at $24^{th}$ and $27^{th}$ epoch.

\begin{table}[htbp]
\footnotesize
\begin{center}
\scalebox{0.9}{
\begin{tabular}{|>{\centering\arraybackslash}m{1.2cm}| >{\centering\arraybackslash}m{1.6cm}|
>{\centering\arraybackslash}m{0.9cm}|c|c|c|}
\hline
Detector & Backbone & Setting & Baseline & LabelEnc & Ours\\
\hline
\multirow{6}{*}{\centering FRCN} & \multirow{4}{*}{\centering R-50} & $1\times$ $ss$ & 37.6 & 38.1 & \textbf{38.3}\\
& & $1\times ms$ & 37.9 & 38.4 & \textbf{38.6}\\
& & $2\times ss$ & 38.0 & 38.9 & \textbf{39.2}\\
& & $2\times ms$ & 39.6 & 39.6 & \textbf{40.4}\\
\cline{2-6}
& R-101 & $2\times ms$ & 41.7 & 41.4 & \textbf{42.3}\\
\cline{2-6}
& R-101 DCN & $2\times ms$ & 44.1 & 44.0 & \textbf{44.9}\\
\hline
\multirow{6}{*}{RetinaNet} & \multirow{4}{*}{\centering R-50} & $1\times ss$ & 36.6 & 37.8 & \textbf{38.3} \\
& & $1\times ms$ & 37.4 & \textbf{38.5} & \textbf{38.5}\\
& & $2\times ss$ & 36.2 & \textbf{39.0} & \textbf{39.0} \\
& & $2\times ms$ & 38.8 & 39.6 & \textbf{40.3} \\
\cline{2-6}
& R-101 & $2\times ms$ & 40.6 & 41.5 & \textbf{42.1} \\
\cline{2-6}
& R-101 DCN & $2\times ms$ & 43.1 & 43.5 & \textbf{44.4} \\
\hline
\multirow{6}{*}{FCOS} & \multirow{4}{*}{\centering R-50} & $1\times ss$ & 38.8 & 39.6 & \textbf{39.7} \\
& & $1\times ms$ & 39.4 & 40.0 & \textbf{40.1}\\
& & $2\times ss$ & 38.1 & \textbf{41.0} & 40.9 \\
& & $2\times ms$ & 41.0 & 41.8 & \textbf{42.3} \\
\cline{2-6}
& R-101 & $2\times ms$ & 42.9 & 43.6 & \textbf{44.1} \\
\cline{2-6}
& R-101 DCN & $2\times ms$ & 44.9 & 45.6 & \textbf{46.3} \\
\hline
\end{tabular}
}
\end{center}
\caption{Detailed comparison with previous SOTA.}
\label{tab:main-exp}
\end{table}

\subsection{Comparison to Teacher-free Methods}

\begin{table}[h]
\begin{center}
\begin{tabular}{|l|c|c|c|c|}
\hline
\multirow{2}{*}{Method} & \multicolumn{2}{c|}{RetinaNet} & \multicolumn{2}{c|}{FRCN} \\
\cline{2-5}
& 1$\times$ $ss$ & 1$\times$ $ms$ & 1$\times$ $ss$ & 1$\times$ $ms$\\
\hline
Baseline & 36.6 & 37.4 & 37.6 & 37.9 \\
\hline
DML$\dag$ & 37.0 & 37.4 & 37.6 & 37.9 \\
\hline
tf-KD$\dag$ & -- & -- & 37.5 & 37.8 \\
\hline
BAN$\dag, \spadesuit$ & 36.8 & 38.0 & 37.6 & 38.1\\
\hline
Ours & \textbf{38.3} & \textbf{38.5} & \textbf{38.3} & \textbf{38.6}\\
\hline
\end{tabular}
\end{center}
\caption{Comparison with typical teacher-free methods. $\dag$ denotes our transfer to detection. $\spadesuit$ denotes reporting the $3^{rd}$ generation result in BAN literature which costs 3$\times$ longer training schedules far more than regular 1$\times$. Also, it is undefined for tf-KD to experiment on RetinaNet with focal loss.}
\label{tab:teacher-free}
\end{table}

\subsubsection{Detailed Comparison with State-of-the-Art.}
As shown in Figure~\ref{fig:horizontal_comparison} and Table~\ref{tab:main-exp}, we compare our LGD framework with the baseline and previous teacher-free SOTA, \textit{i.e.}, the LabelEnc\,\cite{hao2020labelenc} regularization method. We verify the efficacy on MS-COCO on three popular detectors:  Faster R-CNN\,\cite{ren2015faster}, RetinaNet\,\cite{lin2017focal} and FCOS\,\cite{tian2019fcos}. Figure~\ref{fig:horizontal_comparison} shows the result trending as student detector grows stronger (longer periods: $1\times\rightarrow2\times$, scale augmentations: $ss\rightarrow ms$ and larger backbones: R-50 $\rightarrow$ R-101 $\rightarrow$ R-101 DCN). Our model compare favorably to or is slightly better than LabelEnc in earlier settings. For RetinaNet or FCOS R-50 at 2$\times ss$ setting, the baseline runs into overfitting while our method tackles that and achieves \textbf{2.8}\% mAP gain. Notably, as the detector setting becomes stronger, the gain of LabelEnc shrinks rapidly while ours still consistently boosts the performance. For Faster R-CNN with R-101 and R-101 DCN, LabelEnc underperforms the baseline (41.4 \textit{vs.} 41.7 and 44.0 \textit{vs.} 44.1). Instead, our method manage to improve and surpasses LabelEnc at around \textbf{1}\% mAP, verifying higher upper limit. Likewise, for RetinaNet and FCOS with R-101 and R-101 DCN, our method could steadily achieve gains of \textbf{1.2}$\sim$\textbf{1.5}\%. Note that in traditional distillation schemes, it remains unknown to find suitable teacher for such strong students.

\subsubsection{Comparison with Typical Methods.}
As aforementioned, teacher-free schemes other than LabelEnc are NOT designed for detection. For surplus concern, we transfer and reimplement typical methods like DML, tf-KD and BAN to detection by substituting their logits distillation with intermediate feature distillation in mainstream detection KD literature (except tf-KD). As shown in Table~\ref{tab:teacher-free}, these methods obtain slight improvement or are even harmful (tf-KD). BAN performs the best among them. It obtains 0.6\% improvement on RetinaNet 1$\times$ $ms$ R-50 at a cost of actual 3$\times$ training periods. However, it fails to generalize to other settings.

\subsection{Comparison with Classical Teacher-based KD}
\begin{table}[h]
\footnotesize
\begin{center}
       \begin{tabular}{|l|c|c|c|c|}
        \hline
        \multirow{2}{*}{Method} & \multirow{2}{*}{Teacher}& \multicolumn{3}{c|}{Student Backbones} \\
        \cline{3-5}
        & & R-50 & R-101 & R-101 DCN \\
        \hline
        Baseline & N/A &  38.8 & 40.6 & 43.1\\
        \hline
        LabelEnc & N/A & 39.6 &  41.5 & 43.5\\
        \hline
        \multirow{2}{*}{FGFI} & R-101 & 39.8 & 40.7 & 42.4 \\
        \cline{2-5}
        & R-101 DCN & \textbf{40.5} & 41.9 & 43.0 \\
        \hline
        Ours & N/A & 40.3 & \textbf{42.1}  & \textbf{44.4} \\
        \hline
       \end{tabular}
       \caption{Results corresponding to Figure~\ref{fig:distill-50-101-dcnv2-retinanet}. Our method is effective for stronger students compared with others.}
       \label{tab:involve_teacher_based_kd}
    \end{center}
\end{table}

We also compare the proposed \textbf{teacher-free} LGD with the classical \textbf{teacher-based} method, FGFI\,\cite{wang2019distilling}. Experiments are conducted on RetinaNet 2$\times$ $ms$ with backbones R-50, 101 and 101 DCN respectively. As shown in Figure~\ref{fig:distill-50-101-dcnv2-retinanet} and Table~\ref{tab:involve_teacher_based_kd}, our framework performs better when student gets stronger. Towards strong detector with R-101 DCN as backbone, LGD is \textbf{0.9}\% and \textbf{1.4}\% superior to LabelEnc and FGFI. The reason why the benefits of FGFI diminish might attribute to lack of much stronger teacher\,\cite{zhang2021improve,yao2021gdetkd}. We believe it is possible that FGFI with larger teacher or other stronger teacher-based detection KD can outperform ours, but such teacher-presumed setting is not the design purpose of our framework.

\subsection{Ablation Studies}
\begin{table}[htbp]
\begin{center}
       \begin{tabular}{|c|c|c|c|c|c|}
        \hline
        Method & AP & APs & APm & APL & $\Delta$AP\\
        \hline
        N/A &  36.6 & 21.2 & 40.4 & 48.1 & -- \\
        \hline
        MLP &  37.9 & 21.5& 41.9& 49.7& +1.3\\
        \hline
        TransEnc & 37.9 & 21.7& 41.6& 50.2& +1.3\\
        \hline
        PointNet &  38.3 & 23.2& 42.0& 50.0& +1.7\\
        \hline
       \end{tabular}
       \caption{Label Encoder Ablation}
       \label{tab:encoder_ablation}
    \end{center}
\end{table}
\subsubsection{Label Encoding.}
In this work, we adopt PointNet\,\cite{qi2017pointnet} as the label encoding module. In fact, other modules are also applicable. We conduct comparisons on three alternations under $2\times ms$ schedule on MS-COCO with RetinaNet based on ResNet-50 backbone. Specifically, we compare PointNet with a MLP only network, and an encoder network composed of $6$ scaled dot-product attention heads \cite{vaswani2017attention}, abbreviated as ``TransEnc''. Similar to the handling we have done upon PointNet, we feed label descriptors into these networks to obtain label embeddings. We respectively input these label embeddings to remaining LGD modules and examine. All variants achieve good results as shown in Table \ref{tab:encoder_ablation}, which demonstrates the robustness of our 
framework. The PointNet we finally adopt is the best among three of them, perhaps owing to its local-global relationship modeling among label descriptors.

\begin{table}[htbp]
\begin{center}
        \begin{tabular}{|c|ccc|}
        \hline
%        \toprule[1.5pt]
        \multirow{2}{*}{Method} &  \multicolumn{1}{c|}{\multirow{2}{*}{Baseline}} &  \multicolumn{2}{c|}{Interaction Query (Ours)} \\
        \cline{3-4}
        & \multicolumn{1}{c|}{\multirow{2}{*}{}} & \multicolumn{1}{c|}{Label} & Student\\
        \hline
        RetinaNet & \multicolumn{1}{c|}{36.6} & \multicolumn{1}{c|}{37.6 (+1.0)} & \textbf{38.3} (+1.7)\\
        \hline
        FRCN & \multicolumn{1}{c|}{37.6} & \multicolumn{1}{c|}{37.8 (+0.2)}   & \textbf{38.3} (+0.7) \\
        \hline
        FCOS & \multicolumn{1}{c|}{38.8} &\multicolumn{1}{c|}{39.6 (+0.8)}   & \textbf{39.7} (+0.9) \\
        \hline
       \end{tabular}
\end{center}
    \caption{Inter-object Relation Adaption ablations with RetinaNet, Faster R-CNN and FCOS with R-50 1$\times$ $ss$. 
    }
    \label{tab:interaction_query}
\end{table}

\subsubsection{Inter-object Relation Adapter.}
As aforementioned in Sec~\ref{subsec:adapter}, the proposed method adopts the student appearance embeddings as queries and label embeddings as keys and values to involve in the guided inter-object relation modeling (here abbreviated as ``Student''). We also experiment with the reverse option that using label embeddings as queries (abbreviated as ``Label''). As shown in Table~\ref{tab:interaction_query}, for RetinaNet and FRCN 1$\times$ $ss$ with R-50 as backbone, the adopted ``student'' mode are 0.7\% and 0.5\% better than ``Label'' mode.
\begin{table}[htbp]
    \footnotesize
    \begin{subtable}[htbp]{0.20\textwidth}
        \centering
        \begin{tabular}{|c|c|c|}
        \hline
        Object & Context & AP \\
        \hline
         &   & 36.6 \\
        \hline
        &  \Checkmark & 36.6\\
        \hline
        \Checkmark& &  38.0\\
        \hline
        \Checkmark&  \Checkmark & \textbf{38.3}\\
        \hline
       \end{tabular}
       \caption{Embedding Participation}
       \label{tab:knowledge_ingredients}
    \end{subtable}
    \hspace{3mm}
     \begin{subtable}[htbp]{0.18\textwidth}
        \centering
        \begin{tabular}{|c|c|c|c|}
        \hline
        Method & Mode & AP\\
        \hline
        \multirow{3}{*}{RetinaNet} 
        & -- & 36.6 \\
        \cline{2-3}
        & unshared & 37.8 \\
        \cline{2-3}
        & shared & \textbf{38.3}\\
        \hline
        \multirow{3}{*}{FRCN} 
        & -- & 37.6 \\
        \cline{2-3}
        & unshared & 37.7\\
        \cline{2-3}
        & shared & \textbf{38.3} \\
        \hline
       \end{tabular}
       \caption{Head sharing choice}
       \label{tab:head_sharing_ablation}
    \end{subtable}
    \caption{Intra-object knowledge adapter ablations.}
     \label{tab:temps}
\end{table}
\subsubsection{Intra-object Knowledge Mapper.}
As specified in Equation~\ref{eq:projection}, the instructive knowledge is dependent on interacted embeddings of both actual objects and virtual context. We ablate their usage in Table~\ref{tab:knowledge_ingredients}. As expected, the context alone is not helpful since mere context provides nothing useful towards object detection. It manages to enhance the performance when combined with object embeddings (+0.3\%). 

\subsubsection{Head Sharing.} Besides, we also examine the head sharing paradigm as shown in Table~\ref{tab:head_sharing_ablation}. Sharing heads between student and instructive representations is consistently better.

\subsection{Training Efficiency}
\label{subsec:efficiency}
\begin{table}[htbp]
\begin{center}
\begin{tabular}{|c|c|c|c|}
\hline
Method & Pre-training & Overall & Method Specific \\
\hline
Baseline & -- & 12.1 & -- \\ 
\hline
FGFI & 17.0 & 35.5 & 23.4 \\
\hline
LabelEnc & 14.9 & 24.5 & 12.4 \\
\hline
Ours & N/A & \textbf{23.5} & \textbf{11.4} \\
\hline
\end{tabular}
\end{center}
\caption{Comparison of Training Cost (hours).}
\label{tab:train_cost}
\end{table}

Though all distillation and regularization methods won't affect the inference speed of student, they could be training-inefficient due to prerequisite pretraining and distillation process. This is concerned in practical applications but is seldom discussed. As shown in Table~\ref{tab:train_cost}, we benchmark the (1) ``Overall'': overall training cost and (2) ``Method Specific'': overall except student learning (an inherent part shared by all methods). The examination is run on 8 Tesla V100 GPUs upon RetinaNet 2$\times$ $ss$ R-50. We use the corresponding detector with R-101 backbone as teacher for FGFI. Compared with FGFI, we save \textbf{34}\% (23.5 \textit{vs.} 35.5 hours) and \textbf{51}\% (11.4 \textit{vs.} 23.4 hours) on overall and method-specific items respectively. In fact, there could be stronger teacher exploitation for FGFI or other modern teacher-based KDs that outperform ours but it might bring about a heavier training burden and is beyond our discussion scope. Analogous to FGFI, LabelEnc introduces a two-stage training paradigm albeit without pretrained teacher. Towards LabelEnc, our method consumes 1 hour less and is trained in one-step fashion. In practice, LabelEnc consumes $3.8$ G extra GPU footprints except that of the inherent detector, while ours consumes $2.5$ G extra (saving \textbf{34}\% relatively) yet performs better.

\subsection{Versatility}
\subsubsection{Extended Datasets} \hfill\\
\begin{table}[htbp]
\begin{center}
\begin{tabular}{|l|c|c|c|}
\hline
        Method & AP & AP50 & AP75 \\
        \hline
        FRCN & 55.1 & 81.9 & 61.0 \\
        +ours & 56.8 (+\textbf{1.7}) & 82.5 (+0.6) & 63.3 (+\textbf{2.3}) \\
        \hline
        RetinaNet & 56.6 & 81.4 & 61.3  \\
        +ours & 58.9 (+\textbf{2.3}) & 82.6 (+1.2) & 64.3 (+\textbf{3.0})  \\
        \hline
\end{tabular}
\end{center}
\caption{Pascal VOC.}
\label{tab:voc}
\end{table}
\noindent\textbf{(a)} \textit{\textbf{Pascal VOC}}: We conduct experiments with Faster R-CNN and RetinaNet with R-50 under 2$\times$ $ms$ setting. As shown in Table~\ref{tab:voc}, our method improves the results by \textbf{1.7}\% (Faster R-CNN) and \textbf{2.3}\% (RetinaNet). Notably, the AP75 metric of RetinaNet improves \textbf{3.0}\%, showing the efficacy. \\

\begin{table}[htbp]
\begin{center}
\begin{tabular}{|l|c|c|}
\hline
        \diagbox{Method}{mMR}{Detector} & RetinaNet & FRCN \\
        \hline
        Baseline & 57.9 & 48.7\\
        \hline
        Ours & 56.4 ($\uparrow$ \textbf{1.5}) & 46.4 ($\uparrow$ \textbf{2.3}) \\
        \hline
\end{tabular}
\end{center}
\caption{CrowdHuman. mMR: the lower, the better.}
\label{tab:crowds}
\end{table}

\noindent\textbf{(b)} \textit{\textbf{CrowdHuman}}: We also verify our method on the largest crowded detection dataset, CrowdHuman. As shown in Table~\ref{tab:crowds}, our method significantly improves the mMR (lower is better) by \textbf{2.3}\% and \textbf{1.5}\% for Faster R-CNN and RetinaNet respectively. It further demonstrates the generality of our proposed LGD method towards real-world applications.
\subsubsection{Instance Segmentation.}
\begin{table}[htbp]
\begin{center}
\begin{tabular}{|l|c|c|}
\hline
Method  & $\mathrm{AP}_{box}$ & $\mathrm{AP}_{mask}$ \\
\hline
Mask R-CNN (R-50) & 38.8 & 35.2 \\
\, +ours & 39.8 (\textbf{+1.0}) & 36.2 (\textbf{+1.0})\\
\hline
Mask R-CNN (R-101) & 41.2 & 37.2\\
\, +ours & 42.0 (\textbf{+0.8}) & 38.0 (\textbf{+0.8})\\
\hline
\end{tabular}
\end{center}
\caption{Comparison on instance segmentation. }
\label{tab:ins_seg}
\end{table}
To further validate the versatility, we conduct experiments on instance segmentation in MS-COCO. In this task, a detector not only needs to localize each instance but also needs to predict a fine-grained foreground mask. We experiment on Mask R-CNN \cite{he2017mask}. To fully utilize the labels, we replace the object-wise box masks (Section~\ref{subsec:encoder} (2)) with the segmentation masks as better spatial prior. As shown in Table \ref{tab:ins_seg}, our method boosts \textbf{1}\% and \textbf{0.8}\% mask-box AP with respect to Mask R-CNN R-50 and 101.

\section{Conclusion}
In this paper, we propose a brand new self-distillation framework, termed LGD for knowledge distillation in general object detection. It absorbs the spirits of inter-and-intra object relationship into forming the instructive knowledge given regular labels and student representations. The proposed LGD runs in an online manner with decent performance and relatively lower training cost. It is superior to previous teacher-free methods and a classical teacher-based KD method especially for strong student detectors, showing higher potential. We hope LGD could serve as a baseline for future detection KD methods without pretrained teacher. %Codes and logs will be made publicly available in the future.

\section*{Acknowledgements}
This paper is supported by the National Key R\&D Plan of the Ministry of Science and Technology (Project No. 2020AAA0104400) and Beijing Academy of Artificial Intelligence (BAAI). Also, the authors would like to thank Yuxuan Cai and Xiangwen Kong for the proof reading.

%\clearpage

\appendix

\section*{Appendices}
\label{appendix:distill_loss_weight}

\section{Distillation Loss Weight}
The proposed method involves only the distillation loss coefficient $\lambda$ in Eq.~\ref{eq:tot_loss} as a hyperparameter. Actually, we used exponential search with weights $2^{\{-2,-1,0,1,2\}}$ on RetinaNet 1$\times ss$ and obtained 37.3, 37.6, 38.3, 37.9, 37.0. We thus simply adopt weight $\lambda=2^0=1$ throughout all experiments. %and omit it in Eq. 5 \& 7 for simplicity.
%and are welcome for examination of log files in open-source version. 
%(Closer weights like 0.9 and 1.1 perform comparable to 1).

\section{Designs in Label-appearance Encoder}
\textbf{(a) Why LayerNorm not BatchNorm in label encoder}: Despite low dimensionality, the processing of label descriptors are confined with their belonged images. Unlike classification, images in detection datasets like COCO are high-resolution, leading to the most common 2-images-per-GPU protocol. Using BatchNorm provides poor statistics and is 0.5\% inferior to LN (37.8\% \textit{v.s.} 38.3\%) on RetinaNet 1$\times ss$. 

%\subsection{Mask pooling in Appearance Encoding}
\noindent \textbf{(b) Mask pooling in appearance encoder}: We also examined the performance by RoIAlign which performs comparably. We thus adopt the mask pooling which is simpler.

\section{Results on Common 3$\times$ Schedule}
\begin{table}[htbp]
\begin{center}
\begin{tabular}{|c|c|c|c|c|}
\hline
Detector & Backbone & Baseline & Ours & $\Delta$AP \\
\hline
\multirow{2}{*}{\centering RetinaNet} & R-50 & 38.7 & \textbf{40.5} & +1.8 \\
& R-101 & 40.4 & \textbf{42.1} & +1.7\\
\hline
\multirow{3}{*}{\centering FRCN} & R-50 & 40.2 & \textbf{40.9} & +0.7 \\
& R-101 & 42.0 & \textbf{42.7} & +0.7 \\
& X-101 & 43.0 & \textbf{44.1} & +1.1 \\
\hline
\end{tabular}
\end{center}
\caption{Results on 3$\times$ $ms$ setting.}
\label{tab:3xms_expr}
\end{table}
Above experimental sections mainly involve 2$\times$ setting to fairly compare with previous methods in detection KD literature. Here we also release the results on commonly-trained $3\times$ $ms$ settings in general detection literature. Experiments are conducted upon Faster R-CNN and RetinaNet. As shown in Table~\ref{tab:3xms_expr}, the proposed LGD is still robust towards improvement. Besides, Table~\ref{tab:other_detectors} and Table~\ref{tab:other_backbones_swint} also exhibit experimental terms on 3$\times$ $ms$ (POTO and Swin-T variants).

\section{Results on Other Student Variants}

\noindent \textbf{(a) Backbones}: Besides strong student backbones as R-101 and R-101 DCN, we further study the effect on even larger backbones like ResNeXt 101\,\cite{xie2017aggregated,zhu2019deformable} with deformable convolutions v2 (abbreviated as X-101 DCN) and Swin Transformer\,\cite{liu2021swin}. Due to limited resources, for Swin Transformer, we exploit the tiny version of architecture (Swin-T) for experiments.

\begin{table}[htbp]
\begin{center}
\begin{tabular}{|c|c|c|c|c|}
\hline
Detector & Schedule & Baseline & Ours & $\Delta$AP \\
\hline
RetinaNet & \multirow{3}{*}{2$\times$ $ms$} & 43.8 & \textbf{45.9} & +2.1 \\
FRCN & & 45.1 & \textbf{46.1} & +1.0 \\
FCOS & & 46.1 & \textbf{47.9} & +1.8 \\
\hline
\end{tabular}
\end{center}
\caption{Results on X-101 DCN as backbone: For LGD upon the X-101 DCN-based experiments, we early stop the distillation (Eq.~\ref{eq:mse}) at $130k^{th}$ iteration to avoid over-fitting.}
\label{tab:other_backbones_x101dcnv2}
\end{table}

\begin{table}[htbp]
\begin{center}
\footnotesize
\scalebox{0.95}{
\begin{tabular}{|>{\centering\arraybackslash}m{1.2cm}| >{\centering\arraybackslash}m{1.0cm}
|cc|cc|}
\hline
\multirow{2}{*}{Detector} & \multirow{2}{*}{Schedule} & \multicolumn{2}{c|}{Baseline} & \multicolumn{2}{c|}{Ours} \\
\cline{3-6}
& & \multicolumn{1}{c|}{$\mathrm{AP}_{box}$} &  \multicolumn{1}{c|}{$\mathrm{AP}_{mask}$} &   \multicolumn{1}{c|}{$\mathrm{AP}_{box}$} &  \multicolumn{1}{c|}{$\mathrm{AP}_{mask}$} \\ \hline
\multirow{2}{*}{RetinaNet} & \multirow{4}{*}{3$\times$ $ms$} & \multicolumn{1}{c|}{\multirow{2}{*}{44.6}} & \multicolumn{1}{c|}{\multirow{2}{*}{--}} & \multicolumn{1}{c|}{\textbf{46.0}} & \multicolumn{1}{c|}{\multirow{2}{*}{--}} \\
& & \multicolumn{1}{c|}{\multirow{2}{*}{}} & & \multicolumn{1}{c|}{(+1.4)} &  \\
\cline{3-6}
%\multirow{2}{*}{FRCN} &  & \multicolumn{1}{c|}{\multirow{2}{*}{45.1}} & \multicolumn{1}{c|}{\multirow{2}{*}{--}} & \multicolumn{1}{c|}{\textbf{45.7}} & \multicolumn{1}{c|}{\multirow{2}{*}{--}} \\
%& & \multicolumn{1}{c|}{\multirow{2}{*}{}} & & \multicolumn{1}{c|}{(+0.6)} & \\
%\cline{3-6}
\multirow{2}{*}{MRCN} &  & \multicolumn{1}{c|}{\multirow{2}{*}{45.5}} & \multicolumn{1}{c|}{\multirow{2}{*}{41.8}} & \multicolumn{1}{c|}{\textbf{46.4}} & \multicolumn{1}{c|}{\textbf{42.4}}\\

& &\multicolumn{1}{c|}{} & \multicolumn{1}{c|}{}& \multicolumn{1}{c|}{(+0.9)} & \multicolumn{1}{c|}{(+0.6)} \\ 
\hline
\end{tabular}
}
\end{center}
\caption{Results on Swin-T as backbone: We denote by MRCN the Mask R-CNN and report the results of $\mathrm{AP}_{box}$ (\textit{i.e.}, AP) and $\mathrm{AP}_{mask}$, respectively.}
\label{tab:other_backbones_swint}
\end{table}

As shown in Table~\ref{tab:other_backbones_x101dcnv2}, LGD obtains significant improvements of \textbf{2.1}\% and \textbf{1.8}\% on RetinaNet and FCOS with X-101 DCN and also boosts the Faster R-CNN variant by \textbf{1}\%. 

Additionally, the proposed LGD generalizes well to another strong transformer-based backbone Swin-Tiny. As shown in Table~\ref{tab:other_backbones_swint}, RetinaNet with Swin-T backbone is boosted by \textbf{1.4}\%. LGD also improves the Mask R-CNN by 0.9\% and 0.6\% \textit{w.r.t.} the box and mask AP. For simplicity, in experiments upon Swin-T, we basically use the same optimization setting (weight decay, optimizer and learning rate) as that of the Swin-Transformer-based student detectors for LGD modules. We believe that there could be more suitable settings for higher performance by further tuning but we think results by such naive exploitation might be enough to verify the potential. 
We refer to an open-sourced project\,\cite{hu2021swint} based on Detectron2 for implementation of the above Swin-T-based RetinaNet and Mask R-CNN baselines.

%\subsection{Detection Heads}%\section{Results on Other Detectors}
\noindent \textbf{(b) Detection Heads}: Besides typical detection heads as RetinaNet, Faster R-CNN and FCOS referred in the main paper, we also generalize our proposed LGD framework on ATSS\,\cite{zhang2020bridging} and POTO\,\cite{wang2021end}. See Table~\ref{tab:other_detectors} for the details.
\begin{table}[htbp]
\begin{center}
\footnotesize
\scalebox{0.95}{
\begin{tabular}{|>{\centering\arraybackslash}m{0.9cm}| >{\centering\arraybackslash}m{1.6cm}|
>{\centering\arraybackslash}m{1.0cm}|c|c|c|c|}
\hline
Detector & Backbone & Schedule & Baseline & Ours & $\Delta$AP\\
\hline
\multirow{2}{*}{\centering ATSS} & R-50 & $1\times$ $ss$ & 39.3 & \textbf{40.0} & +0.7 \\
& R-101 DCN & $2\times ms$ & 45.8 & \textbf{46.8} & +1.0 \\
\hline
\multirow{2}{*}{\centering POTO} & R-50 & $3\times$ $ms$ & 39.0 & \textbf{41.0} & +2.0 \\
& R-101 & $3\times ms$ & 41.1 & \textbf{42.7} & +1.6 \\
\hline
\end{tabular}
}
\end{center}
\caption{Results on Other Detectors.}
\label{tab:other_detectors}
\end{table}
 
For ATSS, we experiment on the basic R-50 1$\times$ $ss$ then on much stronger setting, dubbed R-101 DCN 2$\times$ $ms$ where ours could improve by \textbf{1}\%, demonstrating the robustness.

For POTO, we report the results on their accustomed R-50 and R-101 3$\times$ $ms$ settings respectively. As could be seen, the proposed LGD framework significantly boosts the performance by \textbf{2.0}\% and \textbf{1.6}\% respectively.

 It is noteworthy that ATSS and POTO introduce different label assign strategies and losses, \textit{e.g.}, adaptive sample selection, Hungarian matching and GIoU loss \cite{rezatofighi2019generalized}. Thus the improvement implies certain orthogonality and demonstrates the generality of our method.

We refer to an open-sourced framework cvpods\,\cite{zhu2020cvpods} for the implementation of these two detectors.

%\section{}
%we have tried transformer decoder that uses student features as query (no appearance encoding and mapper) into the relation adaptater and output instructive feature maps. Whereas, the training is unstable and sometimes falls into trivial solutions. A stable result on RetinaNet 1$\times ss$ is 37.2 (0.6\% better than baseline, but 1.1\% inferior to current mapper practice).

\section{Qualitative Analysis}
\begin{figure*}[htbp]
    \centering
    \begin{subfigure}[t]{0.24\linewidth}
         \centering
         \includegraphics[width=\textwidth]{vis/motor/00031_img_origin.png}
         \label{fig:auxtask1}
     \end{subfigure}
     \begin{subfigure}[t]{0.24\linewidth}
         \centering
         \includegraphics[width=\textwidth]{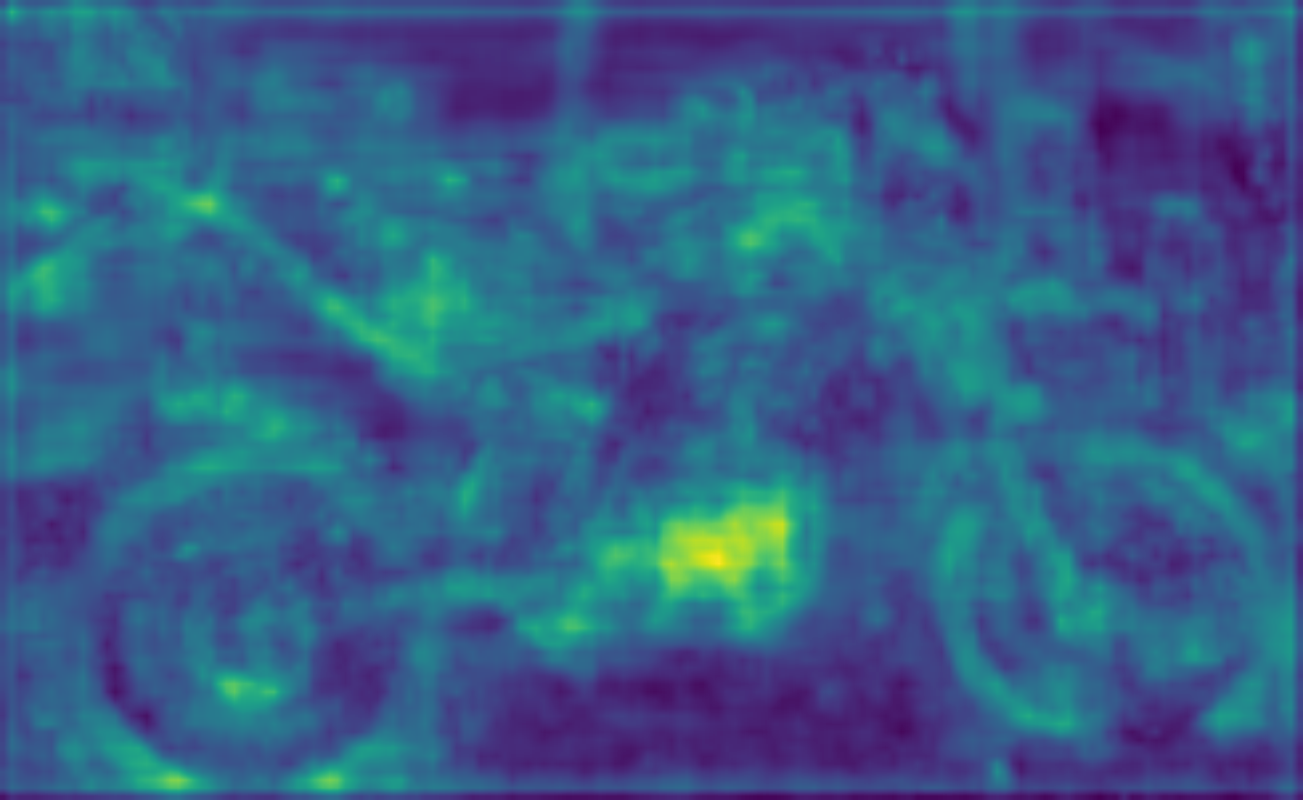}
         \label{fig:auxtask2}
     \end{subfigure}
     \begin{subfigure}[t]{0.24\linewidth}
         \centering
         \includegraphics[width=\textwidth]{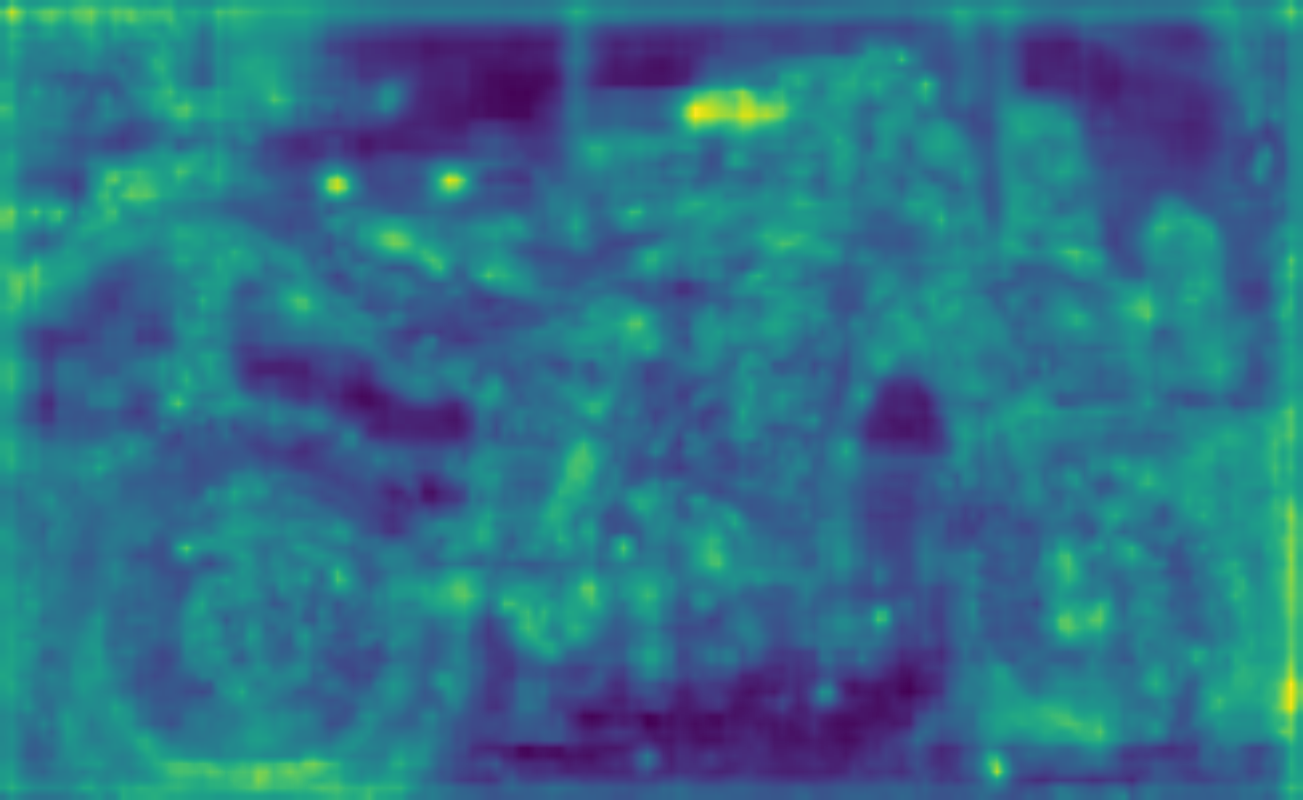}
         \label{fig:auxtask4}
     \end{subfigure}
     \begin{subfigure}[t]{0.24\linewidth}
         \centering
         \includegraphics[width=\textwidth]{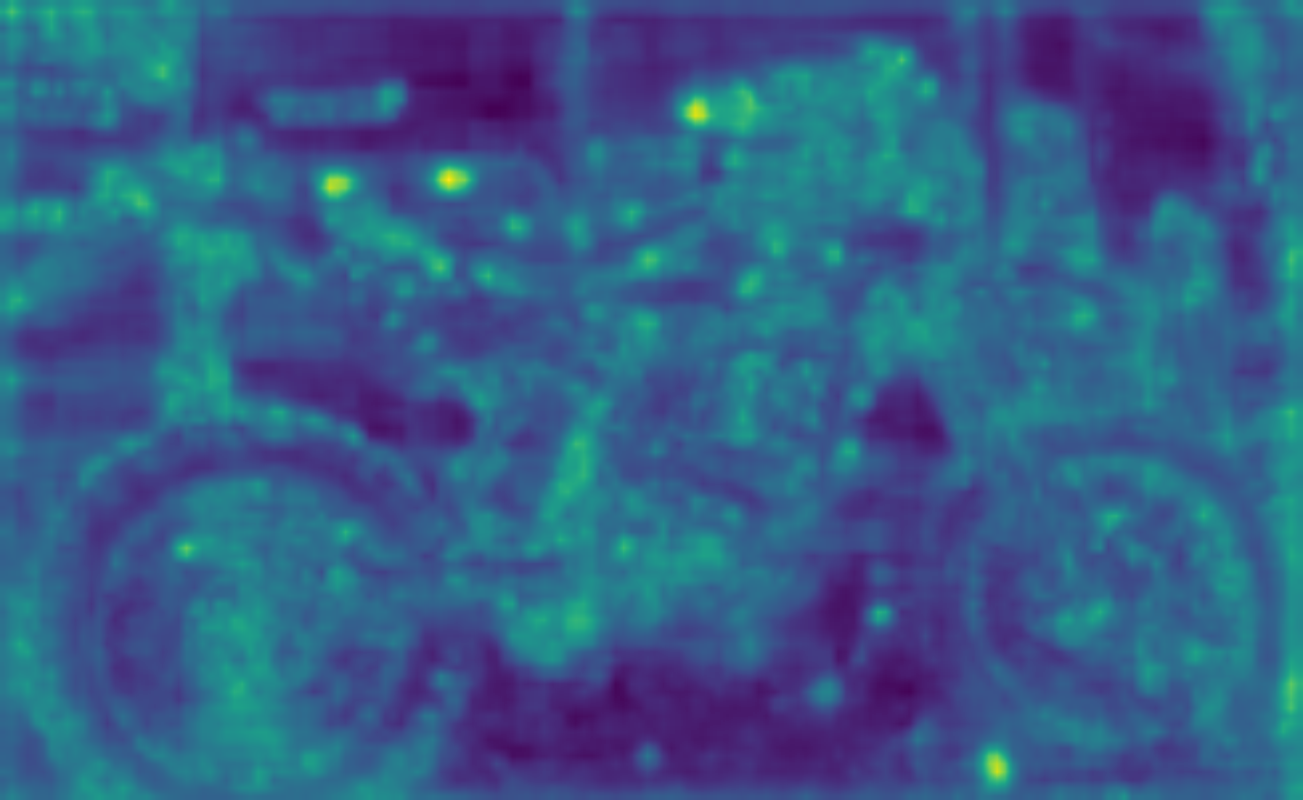}
         \label{fig:auxtask5}
     \end{subfigure}
      \begin{subfigure}[t]{0.24\linewidth}
         \centering
         \includegraphics[width=\textwidth]{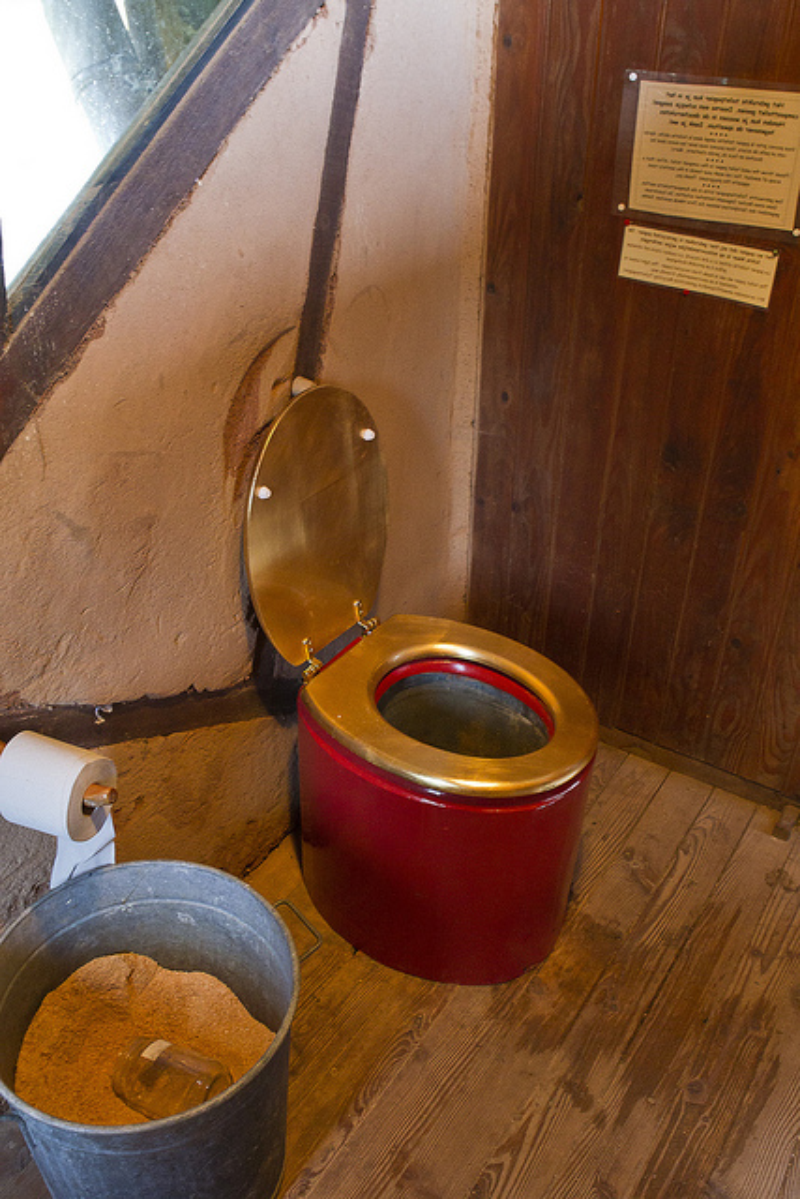}
         \label{fig:auxtask1}
     \end{subfigure}
     \begin{subfigure}[t]{0.24\linewidth}
         \centering
         \includegraphics[width=\textwidth]{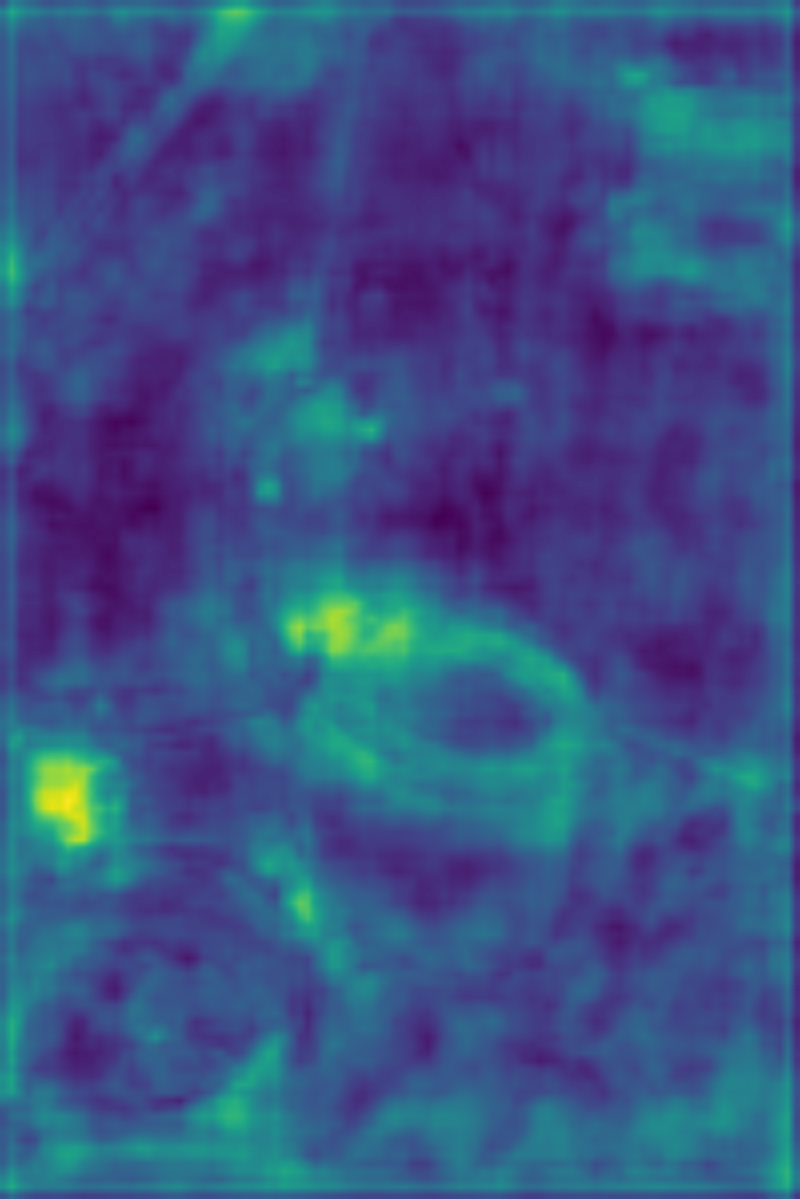}
         \label{fig:auxtask2}
     \end{subfigure}
     \begin{subfigure}[t]{0.24\linewidth}
         \centering
         \includegraphics[width=\textwidth]{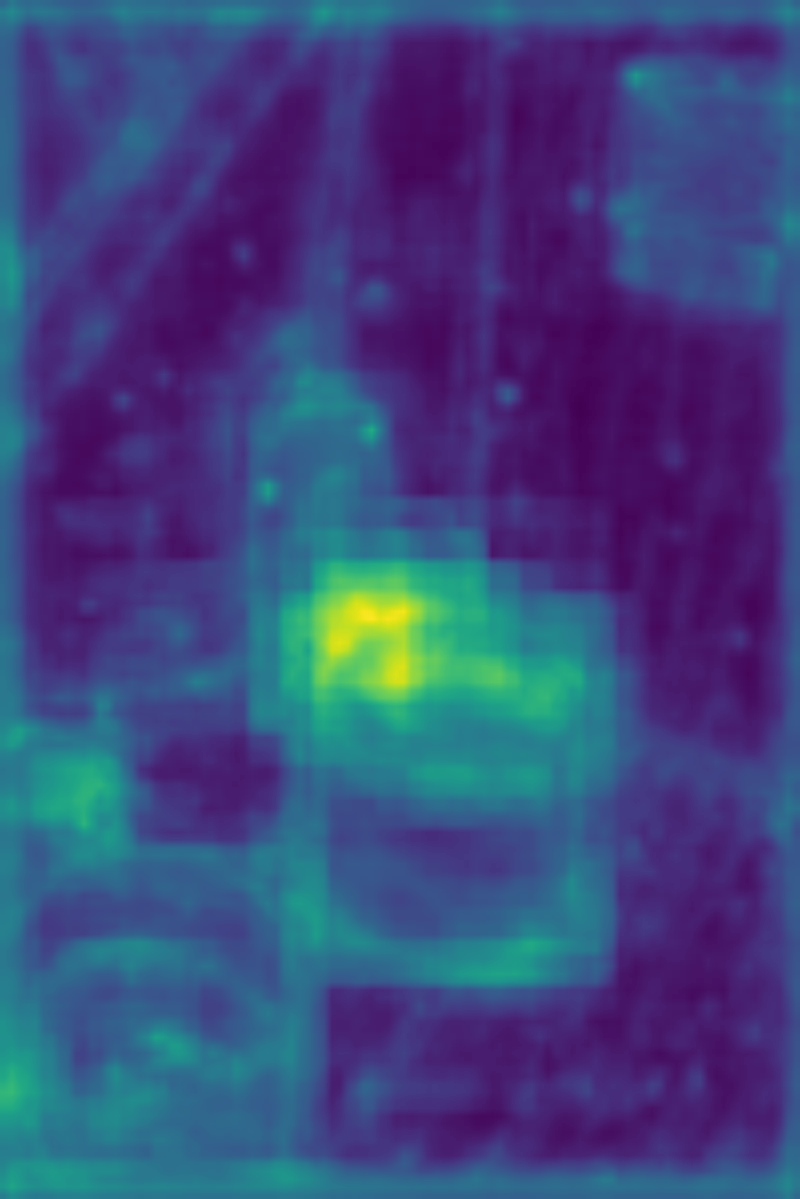}
         \label{fig:auxtask4}
     \end{subfigure}
     \begin{subfigure}[t]{0.24\linewidth}
         \centering
         \includegraphics[width=\textwidth]{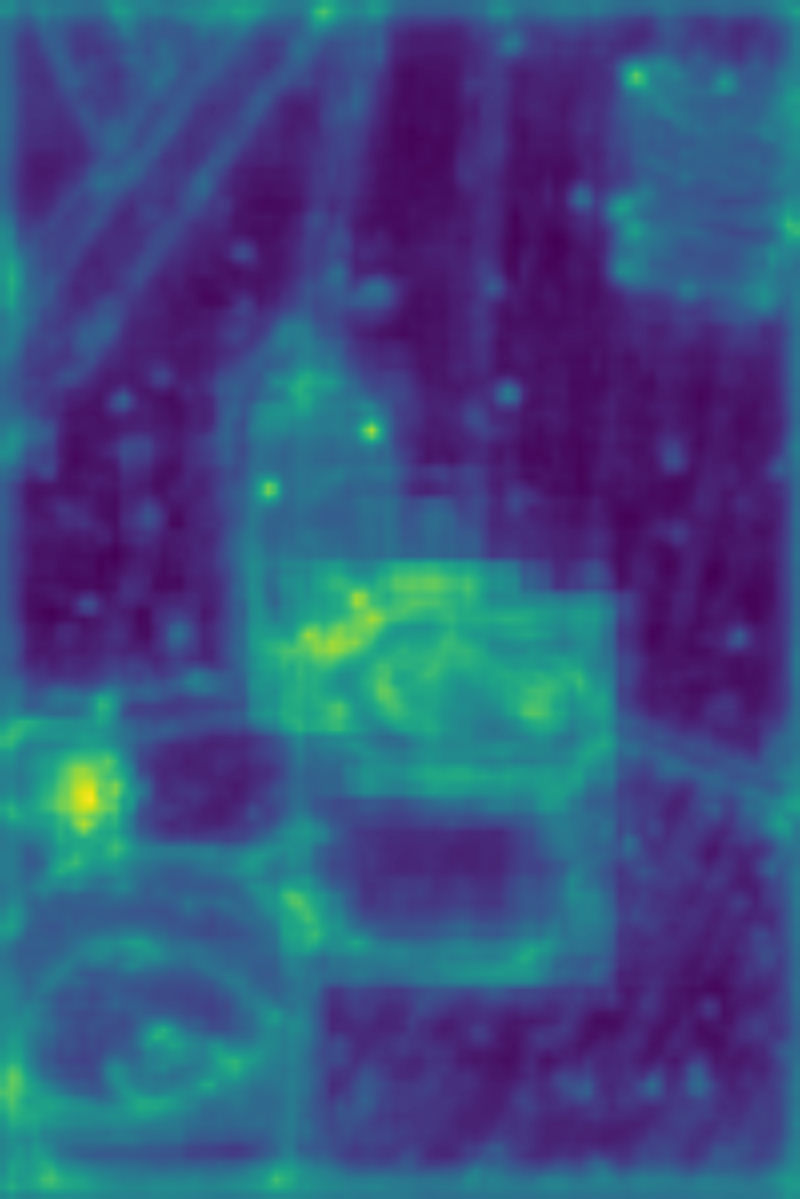}
         \label{fig:auxtask5}
     \end{subfigure}
      \begin{subfigure}[t]{0.24\linewidth}
         \centering
         \includegraphics[width=\textwidth]{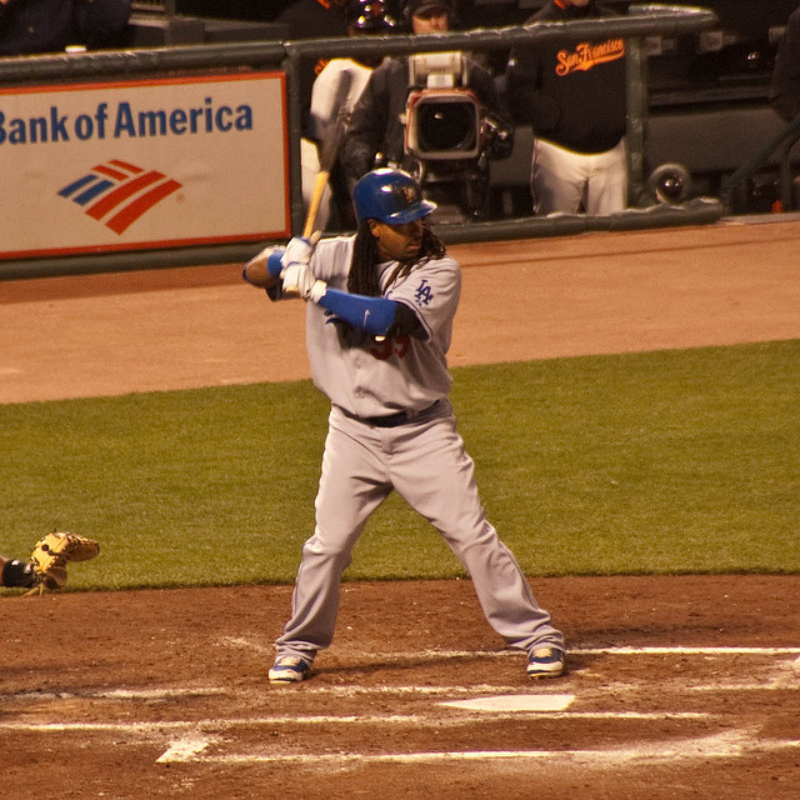}
         \label{fig:auxtask1}
     \end{subfigure}
     \begin{subfigure}[t]{0.24\linewidth}
         \centering
         \includegraphics[width=\textwidth]{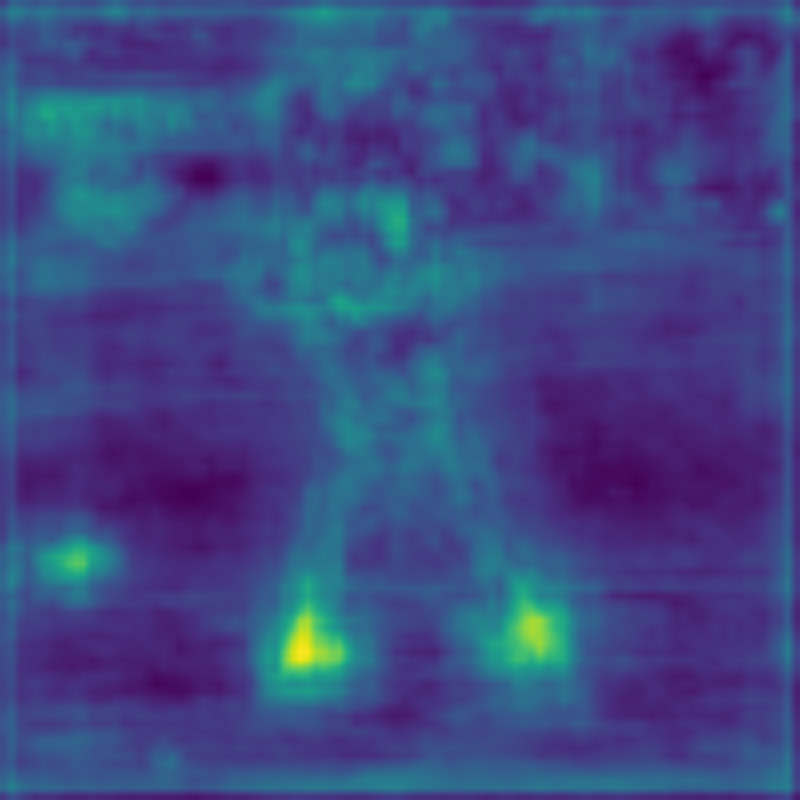}
         \label{fig:auxtask2}
     \end{subfigure}
     \begin{subfigure}[t]{0.24\linewidth}
         \centering
         \includegraphics[width=\textwidth]{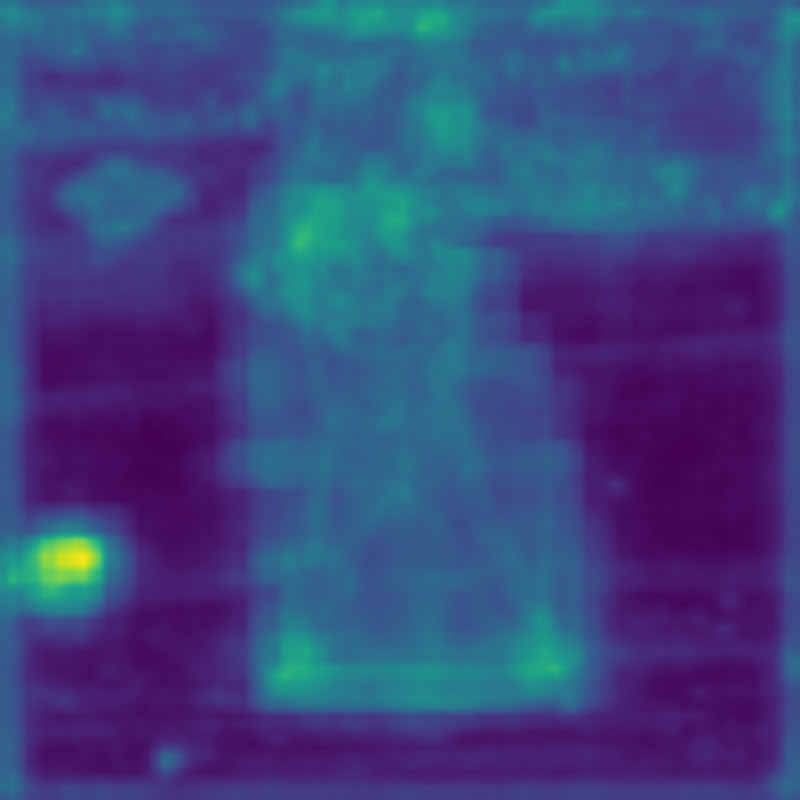}
         \label{fig:auxtask4}
     \end{subfigure}
     \begin{subfigure}[t]{0.24\linewidth}
         \centering
         \includegraphics[width=\textwidth]{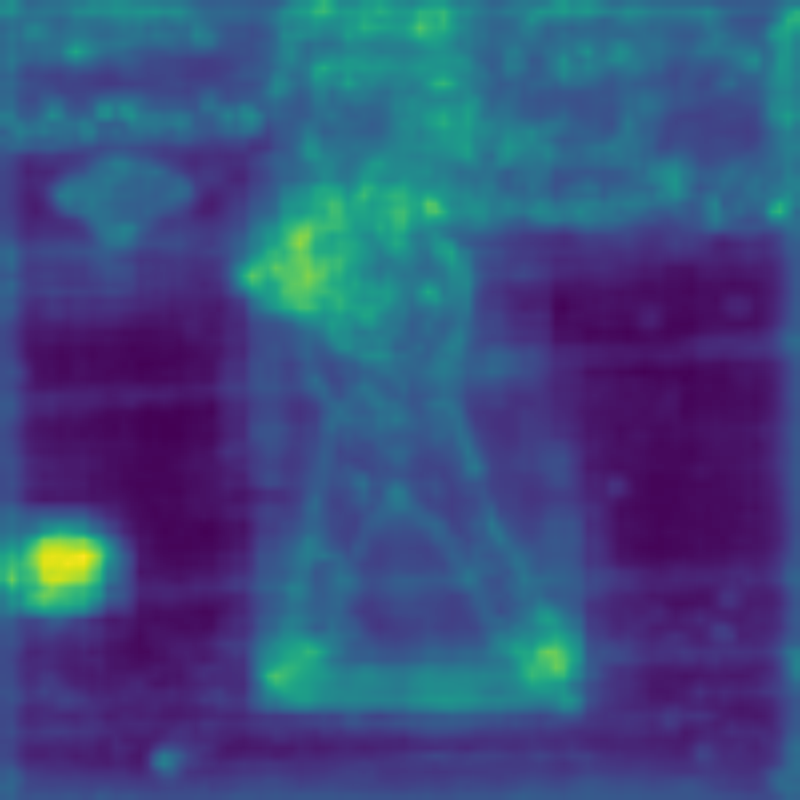}
         \label{fig:auxtask5}
     \end{subfigure}
      \begin{subfigure}[t]{0.24\linewidth}
         \centering
         \includegraphics[width=\textwidth]{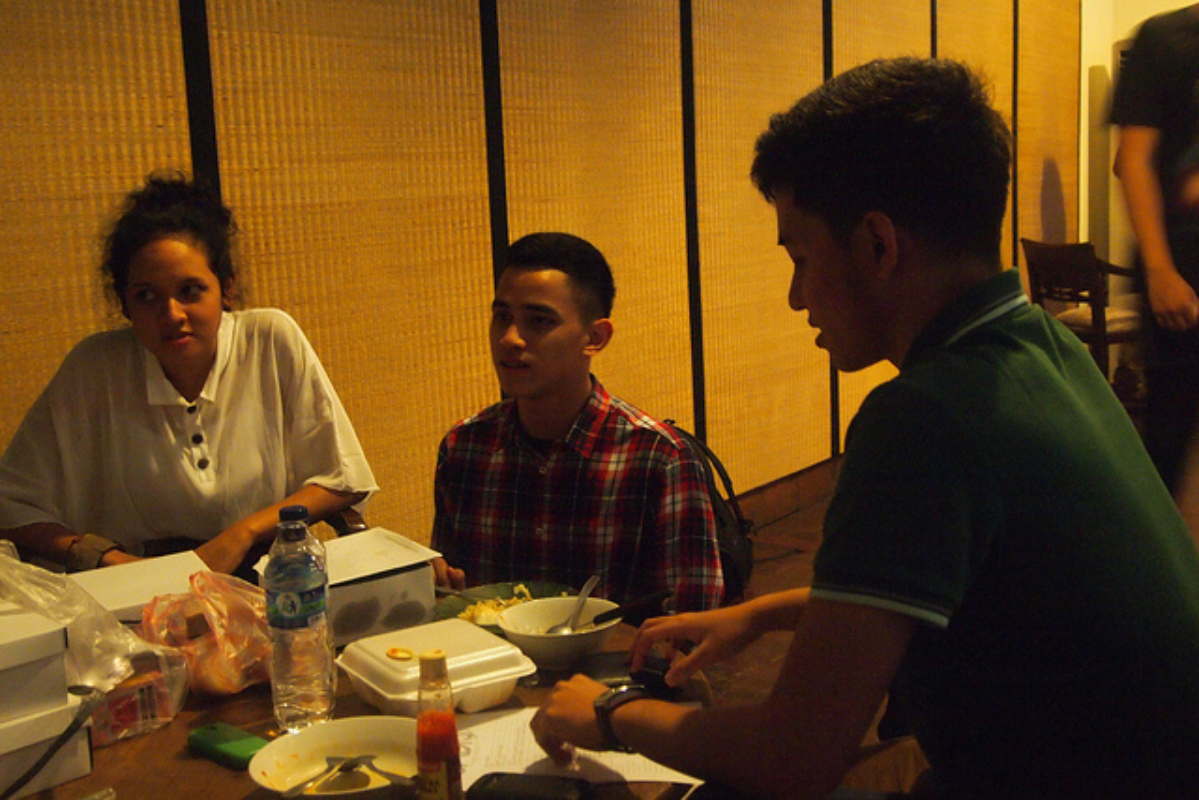}
         \label{fig:auxtask1}
     \end{subfigure}
     \begin{subfigure}[t]{0.24\linewidth}
         \centering
         \includegraphics[width=\textwidth]{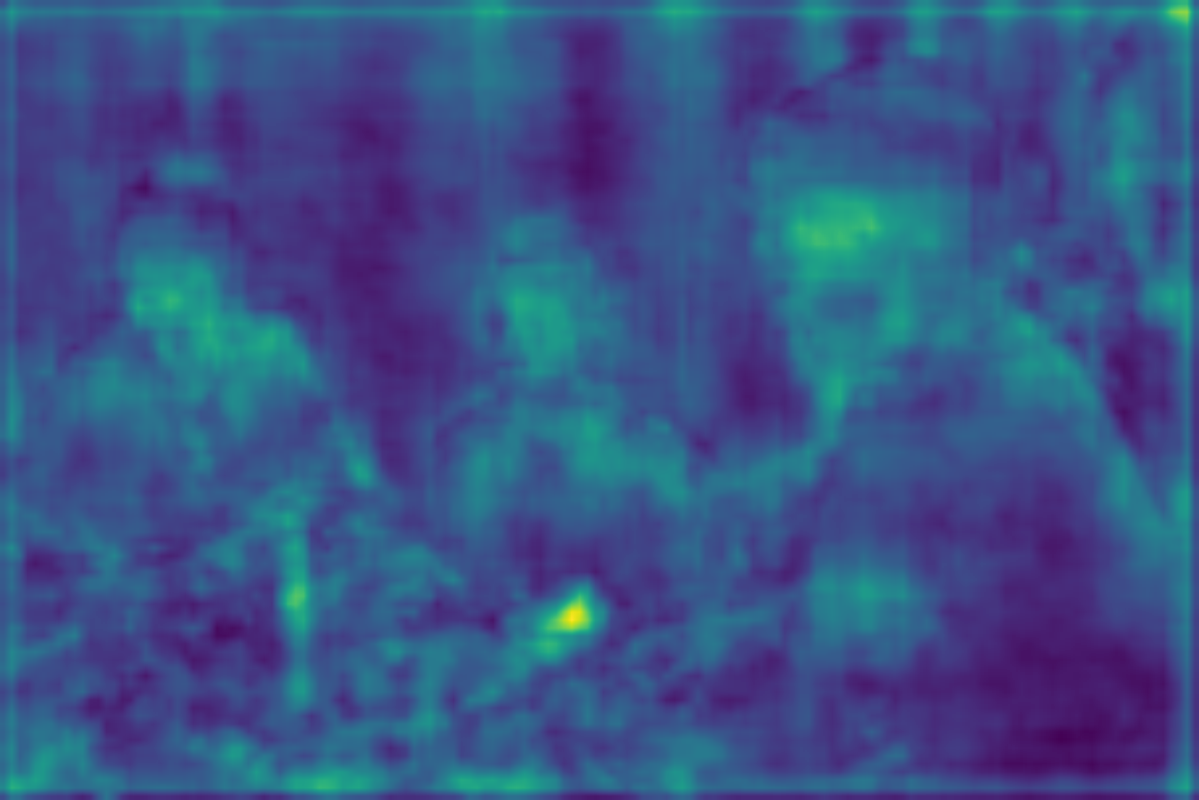}
         \label{fig:auxtask2}
     \end{subfigure}
     \begin{subfigure}[t]{0.24\linewidth}
         \centering
         \includegraphics[width=\textwidth]{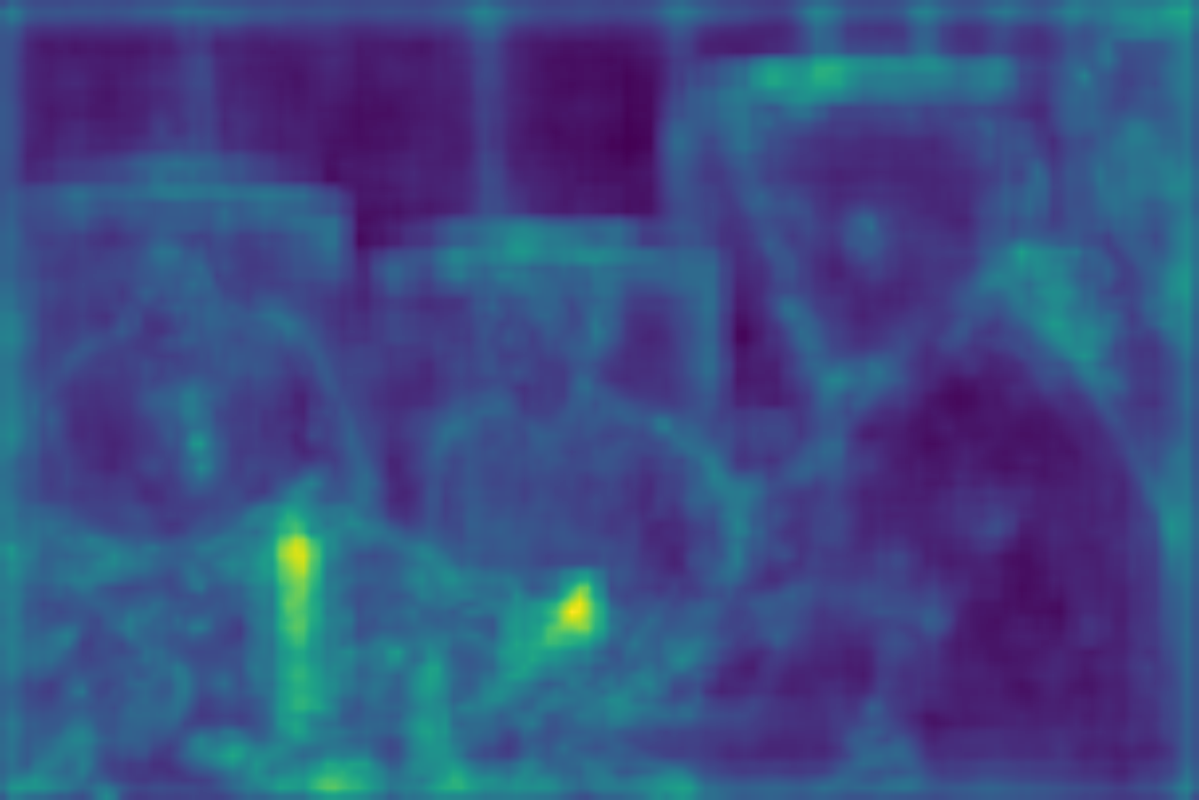}
         \label{fig:auxtask4}
     \end{subfigure}
     \begin{subfigure}[t]{0.24\linewidth}
         \centering
         \includegraphics[width=\textwidth]{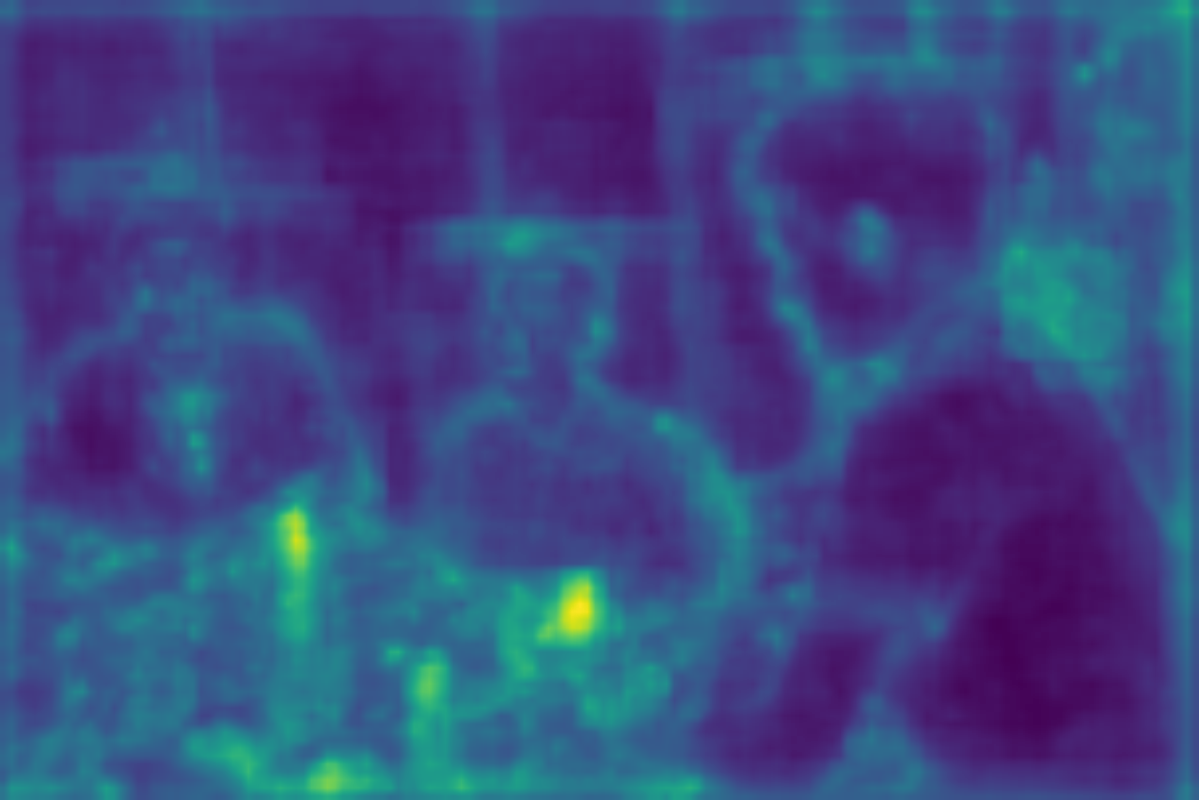}
         \label{fig:auxtask5}
     \end{subfigure}
     \begin{subfigure}[t]{0.24\linewidth}
         \centering
         \includegraphics[width=\textwidth]{vis/tennis/00083_img_origin.png}
         \caption{Input Image}
         \label{fig:auxtask1}
     \end{subfigure}
     \begin{subfigure}[t]{0.24\linewidth}
         \centering
         \includegraphics[width=\textwidth]{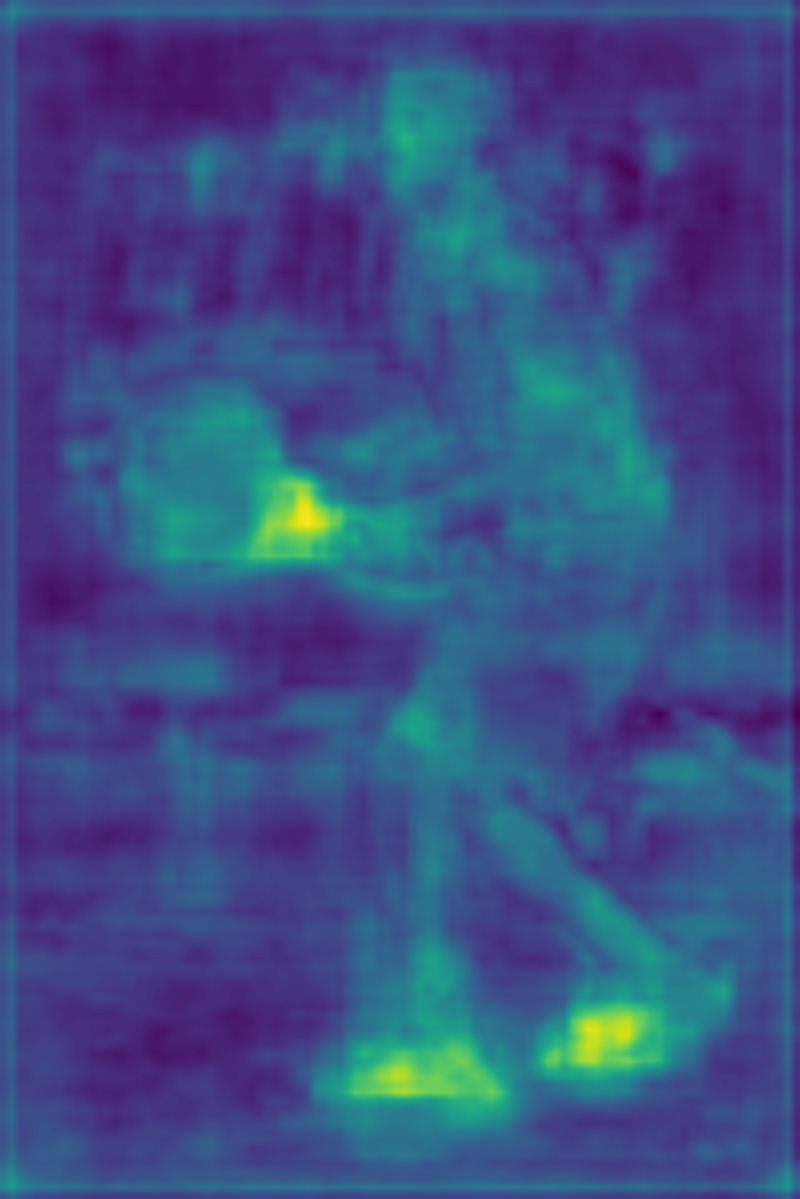}
         \caption{Vanilla Baseline}
         \label{fig:auxtask2}
     \end{subfigure}
     \begin{subfigure}[t]{0.24\linewidth}
         \centering
         \includegraphics[width=\textwidth]{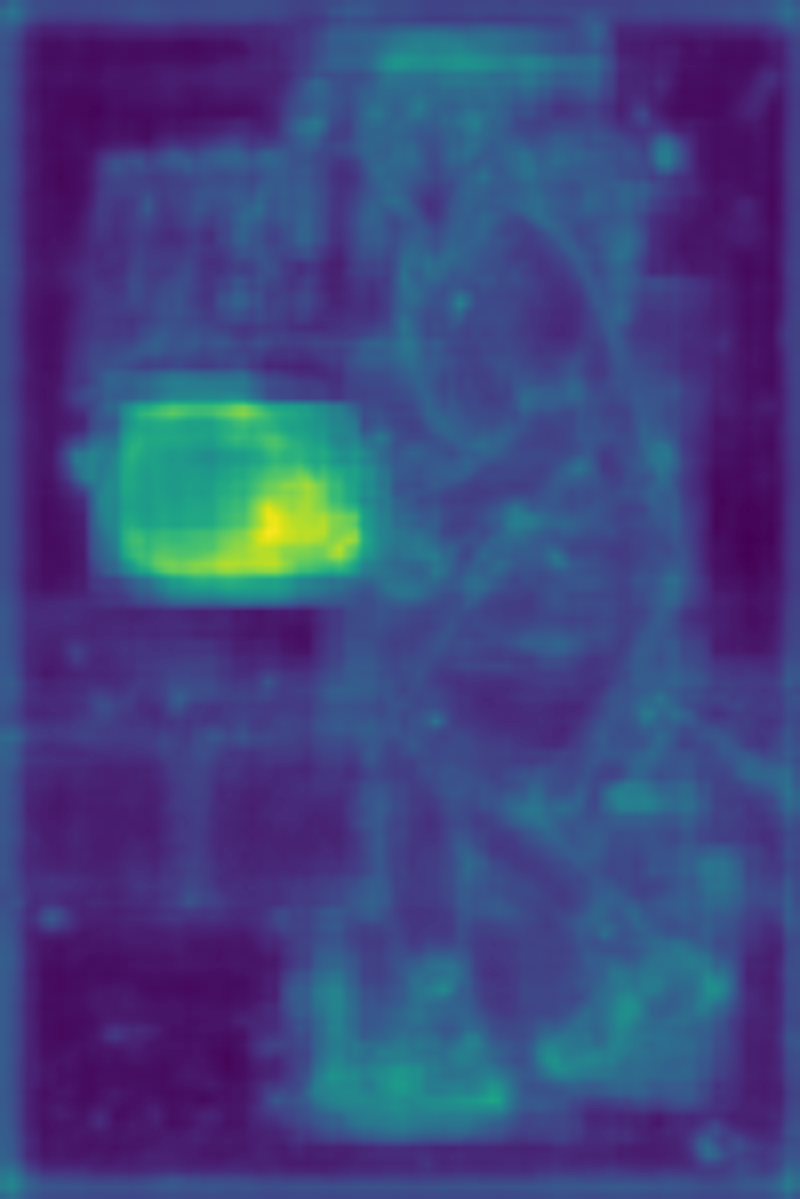}
         \caption{Through LabelEnc}
         \label{fig:auxtask4}
     \end{subfigure}
     \begin{subfigure}[t]{0.24\linewidth}
         \centering
         \includegraphics[width=\textwidth]{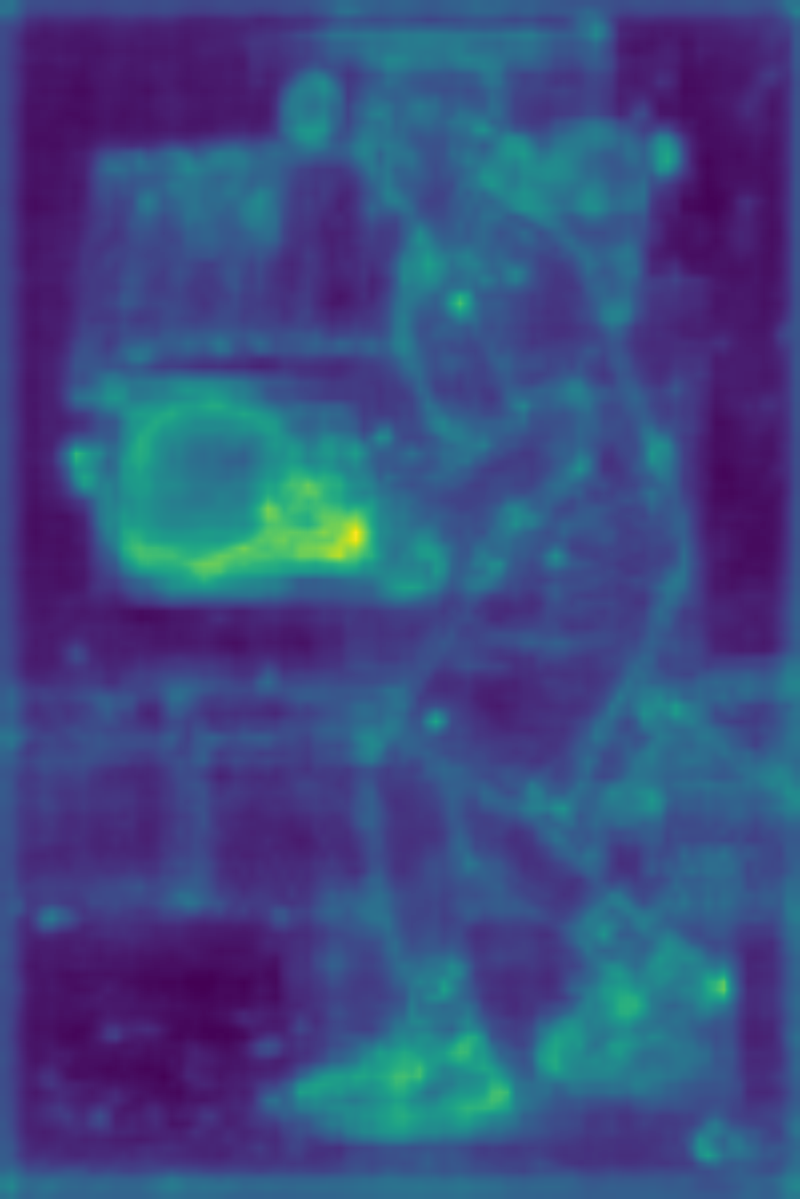}
         \caption{Through Ours}
         \label{fig:auxtask5}
     \end{subfigure}
    \caption{Visualization comparison of the intermediate feature maps. [Best Viewed on Computer]}
  \label{fig:1}
\end{figure*}

Beyond quantitative studies elaborated in the main paper, we briefly analyse the distillation efficacy which reflects certain semantic discrepancy mitigation in a visualized perspective. Specifically, we exhibit the input image and the intermediate student feature maps P3 in feature pyramid by (1) vanilla baseline, (2) LabelEnc regularization and (3) our LGD framework column-by-column. As shown in Figure~\ref{fig:1}, being equipped with our proposed LGD method, the intermediate representations of the student detector not only with much lower background noises but also retains clearer foreground silhouette cues. Rather than highlighting specific local object parts, the feature maps through LGD distillation emphasize on the whole objects that could be more robust towards detection with potential background clutter.

%\nobibliography{aaai22}
\bibliography{aaai22}

\begin{thebibliography}{59}
\providecommand{\natexlab}[1]{#1}

\bibitem[{Ba, Kiros, and Hinton(2016)}]{ba2016layer}
Ba, J.~L.; Kiros, J.~R.; and Hinton, G.~E. 2016.
\newblock Layer normalization.
\newblock \emph{arXiv preprint arXiv:1607.06450}.

\bibitem[{Cai et~al.(2019)Cai, Pan, Ngo, Tian, Duan, and
  Yao}]{cai2019exploring}
Cai, Q.; Pan, Y.; Ngo, C.; Tian, X.; Duan, L.; and Yao, T. 2019.
\newblock Exploring Object Relation in Mean Teacher for Cross-Domain Detection.
\newblock In \emph{CVPR}.

\bibitem[{Chen et~al.(2020{\natexlab{a}})Chen, Mei, Wang, Feng, and
  Chen}]{chen2020online}
Chen, D.; Mei, J.-P.; Wang, C.; Feng, Y.; and Chen, C. 2020{\natexlab{a}}.
\newblock Online knowledge distillation with diverse peers.
\newblock In \emph{AAAI}.

\bibitem[{Chen et~al.(2017)Chen, Choi, Yu, Han, and
  Chandraker}]{chen2017learning}
Chen, G.; Choi, W.; Yu, X.; Han, T.~X.; and Chandraker, M. 2017.
\newblock Learning Efficient Object Detection Models with Knowledge
  Distillation.
\newblock In \emph{NeurIPS}.

\bibitem[{Chen et~al.(2020{\natexlab{b}})Chen, Zhang, Cao, Wang, Lin, and
  Hu}]{chen2020reppointsv2}
Chen, Y.; Zhang, Z.; Cao, Y.; Wang, L.; Lin, S.; and Hu, H. 2020{\natexlab{b}}.
\newblock RepPoints v2: Verification Meets Regression for Object Detection.
\newblock In \emph{NeurIPS}.

\bibitem[{Dai et~al.(2021)Dai, Jiang, Wu, Bao, Wang, Liu, and
  Zhou}]{dai2021general}
Dai, X.; Jiang, Z.; Wu, Z.; Bao, Y.; Wang, Z.; Liu, S.; and Zhou, E. 2021.
\newblock General Instance Distillation for Object Detection.
\newblock In \emph{CVPR}.

\bibitem[{Everingham et~al.(2010)Everingham, Van~Gool, Williams, Winn, and
  Zisserman}]{Everingham10}
Everingham, M.; Van~Gool, L.; Williams, C. K.~I.; Winn, J.; and Zisserman, A.
  2010.
\newblock The Pascal Visual Object Classes (VOC) Challenge.
\newblock \emph{IJCV}.

\bibitem[{Friedman et~al.(2001)Friedman, Hastie, Tibshirani
  et~al.}]{friedman2001elements}
Friedman, J.; Hastie, T.; Tibshirani, R.; et~al. 2001.
\newblock \emph{The elements of statistical learning}, volume~1.
\newblock Springer series in statistics New York.

\bibitem[{Furlanello et~al.(2018)Furlanello, Lipton, Tschannen, Itti, and
  Anandkumar}]{furlanello2018born}
Furlanello, T.; Lipton, Z.~C.; Tschannen, M.; Itti, L.; and Anandkumar, A.
  2018.
\newblock Born-Again Neural Networks.
\newblock In \emph{ICML}.

\bibitem[{Guo et~al.(2021)Guo, Han, Wang, Wu, Chen, Xu, and
  Xu}]{guo2021distilling}
Guo, J.; Han, K.; Wang, Y.; Wu, H.; Chen, X.; Xu, C.; and Xu, C. 2021.
\newblock Distilling Object Detectors via Decoupled Features.
\newblock In \emph{CVPR}.

\bibitem[{Guo et~al.(2020)Guo, Wang, Wu, Yu, Liang, Hu, and
  Luo}]{guo2020online}
Guo, Q.; Wang, X.; Wu, Y.; Yu, Z.; Liang, D.; Hu, X.; and Luo, P. 2020.
\newblock Online Knowledge Distillation via Collaborative Learning.
\newblock In \emph{CVPR}.

\bibitem[{Hao et~al.(2020)Hao, Liu, Zhang, and Sun}]{hao2020labelenc}
Hao, M.; Liu, Y.; Zhang, X.; and Sun, J. 2020.
\newblock LabelEnc: A New Intermediate Supervision Method for Object Detection.
\newblock In \emph{ECCV}.

\bibitem[{He, Girshick, and Doll{\'{a}}r(2019)}]{he2019rethinking}
He, K.; Girshick, R.~B.; and Doll{\'{a}}r, P. 2019.
\newblock Rethinking ImageNet Pre-Training.
\newblock In \emph{ICCV}.

\bibitem[{He et~al.(2017)He, Gkioxari, Doll{\'{a}}r, and Girshick}]{he2017mask}
He, K.; Gkioxari, G.; Doll{\'{a}}r, P.; and Girshick, R.~B. 2017.
\newblock Mask {R-CNN}.
\newblock In \emph{ICCV}.

\bibitem[{Hinton, Vinyals, and Dean(2015)}]{kd}
Hinton, G.; Vinyals, O.; and Dean, J. 2015.
\newblock Distilling the Knowledge in a Neural Network.
\newblock arXiv:1503.02531.

\bibitem[{Hoffman, Gupta, and Darrell(2016)}]{hoffman2016learning}
Hoffman, J.; Gupta, S.; and Darrell, T. 2016.
\newblock Learning with Side Information through Modality Hallucination.
\newblock In \emph{CVPR}.

\bibitem[{Hu et~al.(2018)Hu, Gu, Zhang, Dai, and Wei}]{hu2018relation}
Hu, H.; Gu, J.; Zhang, Z.; Dai, J.; and Wei, Y. 2018.
\newblock Relation Networks for Object Detection.
\newblock In \emph{CVPR}.

\bibitem[{Huang et~al.(2020)Huang, Zou, Kumar, and
  Huang}]{huang2020comprehensive}
Huang, Z.; Zou, Y.; Kumar, B. V. K.~V.; and Huang, D. 2020.
\newblock Comprehensive Attention Self-Distillation for Weakly-Supervised
  Object Detection.
\newblock In \emph{NeurIPS}.

\bibitem[{Ioffe and Szegedy(2015)}]{ioffe2015batch}
Ioffe, S.; and Szegedy, C. 2015.
\newblock Batch Normalization: Accelerating Deep Network Training by Reducing
  Internal Covariate Shift.
\newblock In \emph{ICML}.

\bibitem[{Jaderberg et~al.(2015)Jaderberg, Simonyan, Zisserman, and
  Kavukcuoglu}]{jaderberg2015spatial}
Jaderberg, M.; Simonyan, K.; Zisserman, A.; and Kavukcuoglu, K. 2015.
\newblock Spatial Transformer Networks.
\newblock In \emph{NeurIPS}.

\bibitem[{Kim et~al.(2020)Kim, Ji, Yoon, and Hwang}]{kim2020self}
Kim, K.; Ji, B.; Yoon, D.; and Hwang, S. 2020.
\newblock Self-knowledge distillation: A simple way for better generalization.
\newblock \emph{arXiv preprint arXiv:2006.12000}.

\bibitem[{Lan, Zhu, and Gong(2018)}]{lan2018knowledge}
Lan, X.; Zhu, X.; and Gong, S. 2018.
\newblock Knowledge Distillation by On-the-Fly Native Ensemble.
\newblock In \emph{NeurIPS}.

\bibitem[{Law and Deng(2018)}]{law2018cornernet}
Law, H.; and Deng, J. 2018.
\newblock CornerNet: Detecting Objects as Paired Keypoints.
\newblock In \emph{ECCV}.

\bibitem[{Li, Jin, and Yan(2017)}]{li2017mimicking}
Li, Q.; Jin, S.; and Yan, J. 2017.
\newblock Mimicking Very Efficient Network for Object Detection.
\newblock In \emph{CVPR}.

\bibitem[{Lin et~al.(2017{\natexlab{a}})Lin, Doll{\'{a}}r, Girshick, He,
  Hariharan, and Belongie}]{lin2017feature}
Lin, T.; Doll{\'{a}}r, P.; Girshick, R.~B.; He, K.; Hariharan, B.; and
  Belongie, S.~J. 2017{\natexlab{a}}.
\newblock Feature Pyramid Networks for Object Detection.
\newblock In \emph{CVPR}.

\bibitem[{Lin et~al.(2017{\natexlab{b}})Lin, Goyal, Girshick, He, and
  Doll{\'{a}}r}]{lin2017focal}
Lin, T.; Goyal, P.; Girshick, R.~B.; He, K.; and Doll{\'{a}}r, P.
  2017{\natexlab{b}}.
\newblock Focal Loss for Dense Object Detection.
\newblock In \emph{ICCV}.

\bibitem[{Lin et~al.(2014)Lin, Maire, Belongie, Hays, Perona, Ramanan,
  Doll{\'a}r, and Zitnick}]{lin2014microsoft}
Lin, T.-Y.; Maire, M.; Belongie, S.; Hays, J.; Perona, P.; Ramanan, D.;
  Doll{\'a}r, P.; and Zitnick, C.~L. 2014.
\newblock Microsoft coco: Common objects in context.
\newblock In \emph{ECCV}.

\bibitem[{Liu et~al.(2020)Liu, Rao, Lu, Zhou, and Hsieh}]{liu2020metadistiller}
Liu, B.; Rao, Y.; Lu, J.; Zhou, J.; and Hsieh, C.-J. 2020.
\newblock Metadistiller: Network self-boosting via meta-learned top-down
  distillation.
\newblock In \emph{ECCV}.

\bibitem[{Liu et~al.(2021)Liu, Lin, Cao, Hu, Wei, Zhang, Lin, and
  Guo}]{liu2021swin}
Liu, Z.; Lin, Y.; Cao, Y.; Hu, H.; Wei, Y.; Zhang, Z.; Lin, S.; and Guo, B.
  2021.
\newblock Swin transformer: Hierarchical vision transformer using shifted
  windows.
\newblock \emph{arXiv preprint arXiv:2103.14030}.

\bibitem[{Mostajabi, Maire, and
  Shakhnarovich(2018)}]{mostajabi2018regularizing}
Mostajabi, M.; Maire, M.; and Shakhnarovich, G. 2018.
\newblock Regularizing Deep Networks by Modeling and Predicting Label
  Structure.
\newblock In \emph{CVPR}.

\bibitem[{Park et~al.(2019)Park, Kim, Lu, and Cho}]{park2019relational}
Park, W.; Kim, D.; Lu, Y.; and Cho, M. 2019.
\newblock Relational Knowledge Distillation.
\newblock In \emph{CVPR}.

\bibitem[{Peng et~al.(2020)Peng, Du, Yu, Li, Liao, and Fu}]{peng2020cream}
Peng, H.; Du, H.; Yu, H.; Li, Q.; Liao, J.; and Fu, J. 2020.
\newblock Cream of the crop: Distilling prioritized paths for one-shot neural
  architecture search.
\newblock In \emph{NeurIPS}.

\bibitem[{Qi et~al.(2017)Qi, Su, Mo, and Guibas}]{qi2017pointnet}
Qi, C.~R.; Su, H.; Mo, K.; and Guibas, L.~J. 2017.
\newblock PointNet: Deep Learning on Point Sets for 3D Classification and
  Segmentation.
\newblock In \emph{CVPR}.

\bibitem[{Ren et~al.(2015)Ren, He, Girshick, and Sun}]{ren2015faster}
Ren, S.; He, K.; Girshick, R.~B.; and Sun, J. 2015.
\newblock Faster {R-CNN:} Towards Real-Time Object Detection with Region
  Proposal Networks.
\newblock In \emph{NeurIPS}.

\bibitem[{Rezatofighi et~al.(2019)Rezatofighi, Tsoi, Gwak, Sadeghian, Reid, and
  Savarese}]{rezatofighi2019generalized}
Rezatofighi, H.; Tsoi, N.; Gwak, J.; Sadeghian, A.; Reid, I.; and Savarese, S.
  2019.
\newblock Generalized intersection over union: A metric and a loss for bounding
  box regression.
\newblock In \emph{CVPR}.

\bibitem[{Romero et~al.(2015)Romero, Ballas, Kahou, Chassang, Gatta, and
  Bengio}]{fitnet}
Romero, A.; Ballas, N.; Kahou, S.~E.; Chassang, A.; Gatta, C.; and Bengio, Y.
  2015.
\newblock FitNets: Hints for Thin Deep Nets.
\newblock In \emph{ICLR}.

\bibitem[{Shao et~al.(2018)Shao, Zhao, Li, Xiao, Yu, Zhang, and
  Sun}]{shao2018crowdhuman}
Shao, S.; Zhao, Z.; Li, B.; Xiao, T.; Yu, G.; Zhang, X.; and Sun, J. 2018.
\newblock Crowdhuman: A benchmark for detecting human in a crowd.
\newblock \emph{arXiv preprint arXiv:1805.00123}.

\bibitem[{Szegedy et~al.(2016)Szegedy, Vanhoucke, Ioffe, Shlens, and
  Wojna}]{szegedy2016rethinking}
Szegedy, C.; Vanhoucke, V.; Ioffe, S.; Shlens, J.; and Wojna, Z. 2016.
\newblock Rethinking the Inception Architecture for Computer Vision.
\newblock In \emph{CVPR}.

\bibitem[{Tian et~al.(2019)Tian, Shen, Chen, and He}]{tian2019fcos}
Tian, Z.; Shen, C.; Chen, H.; and He, T. 2019.
\newblock {FCOS:} Fully Convolutional One-Stage Object Detection.
\newblock In \emph{ICCV}.

\bibitem[{Ulyanov, Vedaldi, and Lempitsky(2016)}]{ulyanov2016instance}
Ulyanov, D.; Vedaldi, A.; and Lempitsky, V. 2016.
\newblock Instance normalization: The missing ingredient for fast stylization.
\newblock \emph{arXiv preprint arXiv:1607.08022}.

\bibitem[{Vaswani et~al.(2017)Vaswani, Shazeer, Parmar, Uszkoreit, Jones,
  Gomez, Kaiser, and Polosukhin}]{vaswani2017attention}
Vaswani, A.; Shazeer, N.; Parmar, N.; Uszkoreit, J.; Jones, L.; Gomez, A.~N.;
  Kaiser, L.; and Polosukhin, I. 2017.
\newblock Attention is All you Need.
\newblock In \emph{NeurIPS}.

\bibitem[{Wang et~al.(2021)Wang, Song, Li, Sun, Sun, and Zheng}]{wang2021end}
Wang, J.; Song, L.; Li, Z.; Sun, H.; Sun, J.; and Zheng, N. 2021.
\newblock End-to-end object detection with fully convolutional network.
\newblock In \emph{CVPR}.

\bibitem[{Wang et~al.(2019)Wang, Yuan, Zhang, and Feng}]{wang2019distilling}
Wang, T.; Yuan, L.; Zhang, X.; and Feng, J. 2019.
\newblock Distilling Object Detectors With Fine-Grained Feature Imitation.
\newblock In \emph{CVPR}.

\bibitem[{Wei et~al.(2018)Wei, Pan, Qin, Ouyang, and Yan}]{wei2018quantization}
Wei, Y.; Pan, X.; Qin, H.; Ouyang, W.; and Yan, J. 2018.
\newblock Quantization mimic: Towards very tiny cnn for object detection.
\newblock In \emph{ECCV}.

\bibitem[{Wu et~al.(2019)Wu, Kirillov, Massa, Lo, and
  Girshick}]{wu2019detectron2}
Wu, Y.; Kirillov, A.; Massa, F.; Lo, W.-Y.; and Girshick, R. 2019.
\newblock Detectron2.
\newblock \url{https://github.com/facebookresearch/detectron2}.

\bibitem[{Xie et~al.(2017)Xie, Girshick, Doll{\'{a}}r, Tu, and
  He}]{xie2017aggregated}
Xie, S.; Girshick, R.~B.; Doll{\'{a}}r, P.; Tu, Z.; and He, K. 2017.
\newblock Aggregated Residual Transformations for Deep Neural Networks.
\newblock In \emph{CVPR}.

\bibitem[{Yang et~al.(2019)Yang, Xie, Su, and Yuille}]{yang2019snapshot}
Yang, C.; Xie, L.; Su, C.; and Yuille, A.~L. 2019.
\newblock Snapshot Distillation: Teacher-Student Optimization in One
  Generation.
\newblock In \emph{CVPR}.

\bibitem[{Yang et~al.(2021)Yang, Li, Zhang, Dai, Xiao, Yuan, and
  Gao}]{yang2021focal}
Yang, J.; Li, C.; Zhang, P.; Dai, X.; Xiao, B.; Yuan, L.; and Gao, J. 2021.
\newblock Focal self-attention for local-global interactions in vision
  transformers.
\newblock \emph{arXiv:2107.00641}.

\bibitem[{Yao et~al.(2021)Yao, Pi, Xu, Zhang, Li, and Zhang}]{yao2021gdetkd}
Yao, L.; Pi, R.; Xu, H.; Zhang, W.; Li, Z.; and Zhang, T. 2021.
\newblock G-DetKD: Towards General Distillation Framework for Object Detectors
  via Contrastive and Semantic-guided Feature Imitation.
\newblock arXiv:2108.07482.

\bibitem[{Ye(2021)}]{hu2021swint}
Ye, H. 2021.
\newblock SwinT Reproduction on Detectron2.
\newblock \url{https://github.com/xiaohu2015/SwinT_detectron2}.
\newblock Accessed: 2021-05-18.

\bibitem[{Yuan et~al.(2020)Yuan, Tay, Li, Wang, and Feng}]{yuan2020revisiting}
Yuan, L.; Tay, F. E.~H.; Li, G.; Wang, T.; and Feng, J. 2020.
\newblock Revisiting Knowledge Distillation via Label Smoothing Regularization.
\newblock In \emph{CVPR}.

\bibitem[{Yun et~al.(2020)Yun, Park, Lee, and Shin}]{yun2020regularizing}
Yun, S.; Park, J.; Lee, K.; and Shin, J. 2020.
\newblock Regularizing Class-Wise Predictions via Self-Knowledge Distillation.
\newblock In \emph{CVPR}.

\bibitem[{Zhang and Ma(2021)}]{zhang2021improve}
Zhang, L.; and Ma, K. 2021.
\newblock Improve Object Detection with Feature-based Knowledge Distillation:
  Towards Accurate and Efficient Detectors.
\newblock In \emph{ICLR}.

\bibitem[{Zhang et~al.(2019)Zhang, Song, Gao, Chen, Bao, and
  Ma}]{zhang2019your}
Zhang, L.; Song, J.; Gao, A.; Chen, J.; Bao, C.; and Ma, K. 2019.
\newblock Be Your Own Teacher: Improve the Performance of Convolutional Neural
  Networks via Self Distillation.
\newblock In \emph{ICCV}.

\bibitem[{Zhang et~al.(2020{\natexlab{a}})Zhang, Chi, Yao, Lei, and
  Li}]{zhang2020bridging}
Zhang, S.; Chi, C.; Yao, Y.; Lei, Z.; and Li, S.~Z. 2020{\natexlab{a}}.
\newblock Bridging the Gap Between Anchor-Based and Anchor-Free Detection via
  Adaptive Training Sample Selection.
\newblock In \emph{CVPR}.

\bibitem[{Zhang et~al.(2020{\natexlab{b}})Zhang, Lan, Dai, Zeng, Bai, Chang,
  and Wei}]{zhang2020prime}
Zhang, Y.; Lan, Z.; Dai, Y.; Zeng, F.; Bai, Y.; Chang, J.; and Wei, Y.
  2020{\natexlab{b}}.
\newblock Prime-Aware Adaptive Distillation.
\newblock In \emph{ECCV}.

\bibitem[{Zhang et~al.(2018)Zhang, Xiang, Hospedales, and Lu}]{zhang2018deep}
Zhang, Y.; Xiang, T.; Hospedales, T.~M.; and Lu, H. 2018.
\newblock Deep Mutual Learning.
\newblock In \emph{CVPR}.

\bibitem[{Zhu et~al.(2020)Zhu, Wang, Wang, Yang, Chen, and Li}]{zhu2020cvpods}
Zhu, B.; Wang, F.; Wang, J.; Yang, S.; Chen, J.; and Li, Z. 2020.
\newblock cvpods: All-in-one Toolbox for Computer Vision Research.
\newblock \url{https://github.com/Megvii-BaseDetection/cvpods}.
\newblock Accessed: 2020-12-03.

\bibitem[{Zhu et~al.(2019)Zhu, Hu, Lin, and Dai}]{zhu2019deformable}
Zhu, X.; Hu, H.; Lin, S.; and Dai, J. 2019.
\newblock Deformable ConvNets {V2:} More Deformable, Better Results.
\newblock In \emph{CVPR}.

\end{thebibliography}

\end{document}